\documentclass{article} 
\usepackage[final]{colm2026_conference}

\usepackage{microtype}
\usepackage{hyperref}
\usepackage{url}
\usepackage{booktabs}

\usepackage{inconsolata}
\usepackage{graphicx}

\usepackage[utf8]{inputenc} 
\usepackage[T1]{fontenc}    
\usepackage{hyperref}       
\usepackage{url}            
\usepackage{booktabs}       
\usepackage{amsfonts}       
\usepackage{nicefrac}       
\usepackage{microtype}      
\usepackage[dvipsnames]{xcolor}         

\usepackage{amsmath}
\usepackage{enumitem}
\usepackage{multirow}
\usepackage{graphicx}
\usepackage{wrapfig}
\usepackage{adjustbox}
\usepackage{lineno}
\usepackage{xspace}
\usepackage{makecell}

\usepackage{tcolorbox}
\usepackage{array}

\usepackage{lineno}

\definecolor{darkblue}{rgb}{0, 0, 0.5}
\hypersetup{colorlinks=true, citecolor=darkblue, linkcolor=darkblue, urlcolor=darkblue}

\definecolor{palevioletred}{rgb}{0.86, 0.44, 0.58}
\definecolor{goldenrod}{rgb}{0.85, 0.65, 0.13}
\definecolor{lightblue}{rgb}{0.40, 0.65, 0.75}
\newcommand{\methodname}{\textsc{ScienceMeter}\xspace}

\newcommand{\Sref}[1]{\S\ref{#1}}

\usepackage{fontawesome5}
\usepackage{hyperref}

\title{\methodname:\\ Tracking Scientific Knowledge Updates in Language Models}

\author{Yike Wang \ \ \ Shangbin Feng \ \ \ Yulia Tsvetkov \ \ \ Hannaneh Hajishirzi \\
University of Washington \\
\texttt{yikewang@cs.washington.edu} \\
\href{https://github.com/yikee/ScienceMeter}{\faIcon{github}\, \texttt{https://github.com/yikee/ScienceMeter}}
}

\begin{document}

\ifcolmsubmission
\linenumbers
\fi

\maketitle

\begin{abstract}
Large Language Models (LLMs) are increasingly used to support scientific research, but their knowledge of scientific advancements can quickly  become outdated.
We introduce \methodname, a new framework for evaluating scientific knowledge update methods over scientific knowledge spanning the past, present, and future. \methodname defines three metrics:
\emph{knowledge preservation}, the extent to which models' understanding of previously learned papers is preserved; \emph{knowledge acquisition}, how well scientific claims from newly introduced papers are acquired; and \emph{knowledge projection}, the ability of the updated model to anticipate or generalize to related scientific claims that may emerge in the future. 
Using \methodname, we evaluate the scientific knowledge of LLMs through claim judgment and generation tasks on a curated dataset across ten domains. We evaluate five representative knowledge update approaches and find that the best-performing knowledge update methods can preserve only 85.9\% of existing knowledge, acquire 71.7\% of new knowledge, and project 37.7\% of future knowledge, underscoring that developing robust scientific knowledge update mechanisms is both crucial and challenging.
\end{abstract}

\section{Introduction}
\vspace*{-1mm}

LLMs are being widely used to aid scientific research ~\citep{liang2024mapping, luo2025llm4sr, hsu2024chime, qi2023large, jansen2025discoveryworld}, with the potential to enable even greater future discoveries~\citep{ahn2024transformative, si2024can}. 
However, due to the rapid pace of scientific advancements~\citep{kuhn1962structure} and the static nature of pre-trained LLMs~\citep{bommasani2021opportunities}, their scientific knowledge quickly becomes stale.
We posit that effective scientific knowledge updates in LLMs must do more than add new information, but preserve existing knowledge, incorporate new findings, and enable generalization to reason about future or yet-undiscovered knowledge. 
Although generic update strategies have been explored, e.g., continual pre-training~\citep{gururangan2020don}, instruction-tuning~\citep{wei2021finetuned}, or RAG~\citep{shi2023replug}, it is unclear whether these methods support these goals.  

\begin{figure*}[t]
    \centering
    \includegraphics[width=1.0\textwidth]{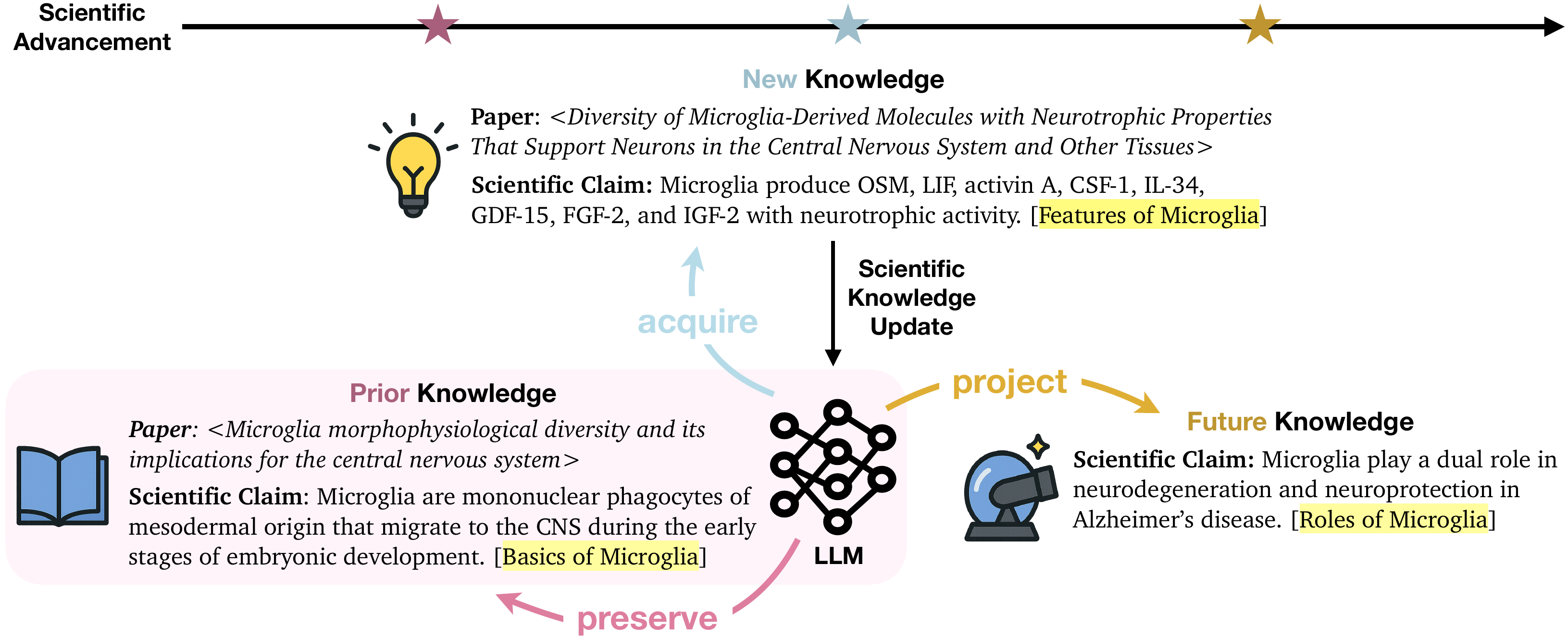}
    \vspace{-6mm}
    \caption{
    We propose an evaluation framework, \methodname, along with novel metrics to quantify the reliability and usefulness of scientific knowledge updates in LLMs: \textcolor{palevioletred}{preservation} of existing scientific claims and their linkage to existing literature, \textcolor{lightblue}{acquisition} of new claims from emerging research, and \textcolor{goldenrod}{projection} of future claims. For example, when we update an LLM with a new paper introducing the features of Microglia, our framework evaluates the acquisition of this new knowledge, as well as the preservation of existing knowledge about the fundamentals of Microglia learned from previous literature, and the ability of LLMs to effectively use its parametric knowledge to extrapolate future knowledge on Microglia, such as its potential roles in Alzheimer’s disease.}
    \vspace{-3mm}
    \label{fig:teaser}
\end{figure*}

To fill this gap, we propose \methodname---a new framework for evaluating how LLMs update and reason over scientific knowledge. 
As shown in Figure~\ref{fig:teaser}, our approach centers on tracking scientific knowledge updates as trajectories along three axes: \emph{preservation} of prior knowledge (the parametric knowledge already encoded in the LLM), \emph{acquisition} of new knowledge introduced through knowledge update methods, and \emph{projection} of future knowledge not yet available to the model but can be inferred. Past discoveries serve as the foundation for future advancements and remain valuable for researchers seeking historical context, validation, or reinterpretation of previous findings, while the latter evaluates the utility of knowledge updates in enabling models to internalize new knowledge, moving beyond mere factual memorization to understand the underlying principles and patterns that govern such knowledge. This capability can facilitate hypothesis generation~\citep{qi2023large} and the formulation of novel ideas~\citep{si2024can}---key future usages of AI for science. 

Inspired by SciFact~\citep{wadden2020fact}, \methodname operationalizes scientific knowledge as \emph{atomic scientific claims}, i.e., atomic verifiable statements expressing a finding about one aspect of a scientific entity or process, which can be verified against a single source. While prior work in general domains often represents knowledge as factoid information or structured entity tuples~\citep{tang2024evowiki, yin2024history}, we argue that scientific claims are more appropriate and meaningful units of knowledge in scientific contexts, as they better capture the core insights and implications of research beyond isolated numerical values. 

In \methodname, we curate a dataset 
spanning ten rapidly evolving scientific fields.
As LLMs become increasingly integrated into these domains, it is essential to evaluate whether existing knowledge update methods can support their progress. 
Related scientific literature is grouped chronologically to represent prior, new, and future knowledge. 
To evaluate scientific knowledge, we focus on two tasks, \emph{claim judgment} and \emph{claim generation}, and emphasize both \emph{factual accuracy} and the model’s \emph{confidence} to better reflect the rigor of the scientific domain.
Specifically, we categorize the model's knowledge of a claim as \textit{correct} (factually accurate and confident), \textit{incorrect} (factually inaccurate and confident), or \textit{unknown} (not confident) and quantify the percentage of two types of errors in preservation, acquisition, and projection, respectively.

We evaluate LLMs' scientific knowledge updates using five methods spanning training, inference, or both. Experimental results across standard and frontier models  highlight that the best-performing knowledge update method achieves on average  only 85.9\% on knowledge preservation, 71.7\% on knowledge acquisition, and 37.7\% on knowledge projection in the domain of \textit{Computer Science}. While inference-time update methods tend to be effective for large models, smaller models require training-based approaches to achieve comparable performance. Cross-domain analysis reveals that performance on these objectives is correlated, with knowledge preservation and projection heavily influenced by domain volatility, while the availability of domain knowledge during pretraining has limited impact. Moreover, even applying on domain-adapted scientific LLMs struggle to balance these objectives, underscoring persistent challenges in updating scientific knowledge in LLMs.

\vspace*{-1mm}
\section{\methodname}
\label{sec:framework}
\vspace*{-1mm}

\begin{figure*}[t]
    \centering
    \includegraphics[width=1.0\textwidth]{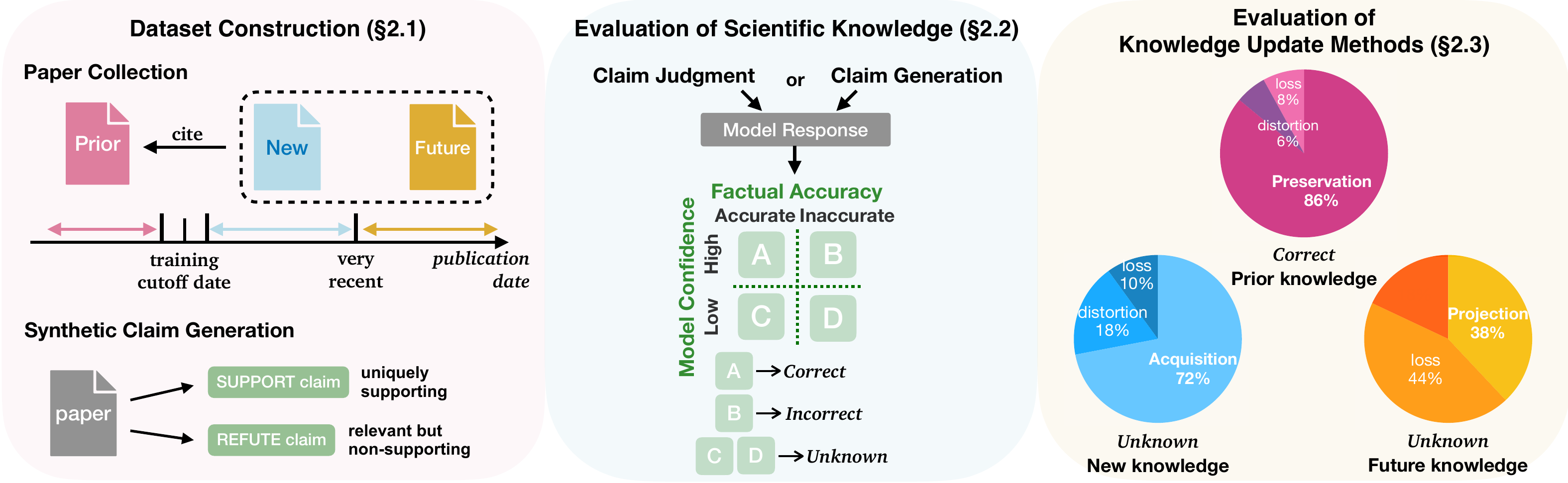}
    \caption{An overview of \methodname: (1) We curate chronologically organized datasets of scientific papers and claims; (2) define claim judgment and generation tasks to evaluate scientific knowledge, incorporating both factual accuracy and model confidence; and (3) introduce metrics for evaluating scientific knowledge updates.}
    \label{fig:overview}
    \vspace*{-4mm}
\end{figure*}

To systematically evaluate scientific knowledge updates in LLMs, \methodname integrates three core components: (1) 
a carefully curated multi-domain dataset of scientific papers and scientific claims (\Sref{sec:data_source_and_data_construction}); (2) evaluation of a model’s scientific knowledge through claim judgment and generation tasks, assessed by both factual accuracy and model confidence (\Sref{sec:evaluation_of_scientific_knowledge}); and (3) novel metrics for evaluating knowledge update methods that aggregate claims from past/present/future data sets (\Sref{sec:evaluation_of_knowledge_update_methods}). An overview of \methodname is in Figure~\ref{fig:overview}.

\subsection{Dataset Construction}
\label{sec:data_source_and_data_construction}

\paragraph{Paper Collection}

We identify 10 rapidly evolving scientific domains, including \emph{Computer Science}, \emph{Medicine}, \emph{Biology}, \emph{Materials Science}, \emph{Psychology}, \emph{Business}, \emph{Political Science}, \emph{Environmental Science}, \emph{Agricultural and Food Sciences}, and \emph{Education}. For each domain, we randomly retrieve 1,000 journal or conference papers (excluding review or survey papers) published at least three months before the knowledge cutoff date of the given model through Semantic Scholar~\citep{ammar2018construction}, an openly accessible academic search platform that indexes publisher-authorized metadata and papers. 
To obtain more recent knowledge on the subjects relevant to each paper, we additionally retrieve papers that cite the original paper (Appendix~\ref{sec:discussion_group_knowledge}) and were published at least three months after the knowledge cutoff date. We also set a recent cutoff date, beyond which papers serve as a proxy for future knowledge. 
In total, we constructed 5,148 triplets of $(p_\text{prior}, p_\text{new}, p_\text{future})$, each representing a prior, new, and future version of scientific knowledge on the same topic (Appendix~\ref{sec:dataset_details}). An additional filtering will be applied to these papers based on the actual model knowledge (Section~\ref{sec:evaluation_of_knowledge_update_methods}).

\paragraph{Synthetic Claim Generation}
We invite \textsc{GPT-4o} to synthetically generate one \textsc{support} (uniquely supporting) scientific claim and one \textsc{refute} (relevant but non-supporting) scientific claim for each paper, resulting in a total of 15,444 $(p, c_\textsc{support})$ and 15,444 $(p, c_\textsc{refute})$ tuples.
Expert evaluation by PhD students in their respective fields shows that, on average, 90.2\% of the generated claims strictly comply with the specified \textsc{support} or \textsc{refute} rules, while 98.7\% broadly align with these rules. 
Further details are provided in Appendix~\ref{sec:dataset_details}.

\subsection{Evaluation of Scientific Knowledge}
\label{sec:evaluation_of_scientific_knowledge}
Now we want to evaluate a model’s scientific knowledge on the collected papers, specifically the scientific claims made in each paper. We propose two tasks, judgment and generation, and categorize the model’s knowledge of a claim as \textit{correct}, \textit{incorrect}, or \textit{unknown}, based on both the response’s \emph{factual accuracy} and the model’s \emph{confidence}.

\paragraph{Task Formulation}

Considering the cases that different papers may present conflicting or opposing claims, for example, one paper may report that drinking water is beneficial for health, while another may argue that excessive water intake can be harmful under certain conditions, we include paper titles in the task setting when evaluating both prior and newly introduced knowledge. This design makes the task more challenging: models must not only understand or memorize individual claims, but also associate claims with their sources and distinguish between potentially conflicting findings across papers.

\begin{itemize}[leftmargin=0.5cm]
\item{\bf Claim Judgment} 
To evaluate a model’s knowledge of $p_\text{prior}$ or $p_\text{new}$, we frame the task as a claim verification problem: given the title $t$ of a prior or new scientific paper and its associated claim $c$, the ground-truth label is $y(c, t) \in \{\textsc{support}, \textsc{refute}\}$. The model is instructed to predict a label $\hat{y}(c, t)$ for each $(c, t)$ pair. 
To evaluate a model’s knowledge of $p_\text{future}$, we adapt the task into a classification setting: given a claim $c$ associated with a ``future'' paper, the ground truth label $y(c) \in \{\textsc{support}, \textsc{refute}\}$ indicates whether its associated ``future'' paper supports or refutes the claim $c$. The model is asked to predict a label $\hat{y}(c)$ based solely on its internalized knowledge or extrapolative reasoning, without access to the actual paper.
\item{\bf Claim Generation} 
The generation task poses a greater challenge. To evaluate a model’s knowledge of $p_\text{prior}$ or $p_\text{new}$, the model is given the title $t$ of a prior or new scientific paper $p$ and instructed to generate a supporting claim $\hat{c}(t)$ such that $y(\hat{c}, t) = \textsc{support}$. 
To evaluate a model’s knowledge of $p_\text{future}$, the model is given the subject $s$ of a ``future'' scientific paper $p$ (with title $t$) and tasked with generating a supporting claim $\hat{c}(s)$ such that $y(\hat{c}, t) = \textsc{support}$.
\end{itemize}


\paragraph{Task Evaluation}

Our evaluation methodology emphasizes both \emph{factual accuracy} and the model’s \emph{confidence}. We present our choices of measurement methods in Section~\ref{sec:factual_correct_and_model_confidence_measurement_methods}. By combining factual accuracy with model confidence, we categorize the model’s knowledge of a claim as \textit{correct} (factually accurate and confident), \textit{incorrect} (factually inaccurate and confident), or \textit{unknown} (not confident). We argue that when confidence is low, even a factually accurate answer is not reliable as it may result from hallucination or random chance. This categorization enables a detailed analysis of the impact on scientific knowledge following knowledge updates, as discussed in the next section.

\subsection{Evaluation of Knowledge Update Methods} 
\label{sec:evaluation_of_knowledge_update_methods}
Given the set of papers and associated claims, along with the model’s knowledge about each claim (\textit{correct}, \textit{incorrect}, or \textit{unknown}), we can systematically evaluate how a given knowledge update method impacts the model’s prior, new, and future scientific knowledge. Specifically, we define three core metrics, Knowledge Preservation, Knowledge Acquisition, and Knowledge Projection, as well as two associated error categories, \texttt{distortion} and \texttt{loss}, as detailed below.

Let $\mathcal{P}_\text{prior}$, $\mathcal{P}_\text{new}$, and $\mathcal{P}_\text{future}$ be sets of prior, new, and future scientific documents in a particular scientific domain. $\mathcal{P}_\text{new}$ is introduced using the given scientific knowledge update method $f$. 
While documents are collected in such a way that $\mathcal{P}_\text{prior}$ is highly likely to appear in the training data of the model, whereas $\mathcal{P}_\text{new}$ and $\mathcal{P}_\text{future}$ are not (as described in Section~\ref{sec:data_source_and_data_construction}), for each task we only take into account scientific claims in $\mathcal{P}_\text{prior}$ that are not \textit{unknown} and claims in $\mathcal{P}_\text{new}$ and $\mathcal{P}_\text{future}$ that are \textit{unknown} to the model before knowledge updates.
Specifically, let $g$ represent either the claim judgment or generation task presented in Section~\ref{sec:evaluation_of_scientific_knowledge}. Then, given a pre-trained language model $\mathrm{LM}$ and a knowledge update method $f$, we define:

\begin{itemize}[leftmargin=0.5cm]
    \item \textbf{Knowledge Preservation} 
    as the percentage of claims in $\mathcal{P}_\text{prior}$ that remain \textit{correct}. The proportion of previously \textit{correct} claims that become \textit{incorrect} is referred to as a \texttt{distortion} in preservation, while the proportion that becomes \textit{unknown} is considered as \texttt{loss}.

    \item \textbf{Knowledge Acquisition} 
    as the proportion of claims in $\mathcal{P}_\text{new}$ that the model $\mathrm{LM}$ correctly acquires through $f$, i.e., changing from \textit{unknown} to \textit{correct}. Similarly, the proportion of \textit{unknown} claims that become \textit{incorrect} is referred to as a \texttt{distortion} in acquisition, while the proportion that stays as \textit{unknown} is referred to as \texttt{loss}.

    \item \textbf{Knowledge Projection} as the percentage of claims in $\mathcal{P}_\text{future}$ that the model $\mathrm{LM}$ successfully projects after update $f$, i.e., claims changing from \textit{unknown} to \textit{correct}. Because some incorrect projections may become correct over time, the true magnitude of Knowledge Projection is likely higher than our current estimate. We define the proportion of \textit{unknown} claims that remain \textit{unknown} as \texttt{loss}. See Appendix~\ref{sec:discussion_knowledge_projection} for further discussion.

\end{itemize}
We provide the detailed formulas for each metric in Appendix~\ref{sec:metrics}. The sum of Knowledge Preservation, \texttt{distortion}, and \texttt{loss} equals one, and the same holds for Acquisition. Among the error types, \texttt{distortion} is considered more severe than \texttt{loss} in both Preservation and Acquisition scenarios. This is because generating a factually inaccurate response with high confidence (e.g., stating ``this claim is \textsc{support} for sure'' to a \textsc{refute} claim) is more problematic than producing a low-confidence response, regardless of its factual accuracy (e.g., ``maybe this claim is \textsc{support}/\textsc{refute}''). An optimal scientific knowledge update method should collectively maximize Knowledge Preservation, Acquisition, and Projection.

\section{Experiment Settings and Results}
\label{sec:experiment_settings_and_results}
In this section, we evaluate five knowledge update methods, covering training, inference, or a combination of both. 
Due to cost constraints, we conduct experiments in the \textit{Computer Science} domain in this section, and present cross-domain analysis in Section~\ref{sec:domains}.
Experiments on both standard and frontier models, along with three confidence measurement approaches, demonstrate the challenge of developing a reliable scientific knowledge update method capable of meeting all three objectives. 

\subsection{Models}
\label{sec:models} 
We aim to evaluate the performance of various knowledge update methods on a widely used, reasonably sized model and a frontier large model.
As a representative of commonly used mid-sized models, we select LLaMA3.1-8B-Instruct~\citep{grattafiori2024llama}, and OLMo2-32B-Instruct~\citep{olmo20242} serves a representative of frontier models, given computational constraints and the limited openness of commercial models. Notably, OLMo2-32B-Instruct has demonstrated frontier performance while requiring only one-third of the compute of other open-weight models and outperforming GPT-4o mini~\citep{openai2024gpt4o}. See Appendix~\ref{sec:scientific_llms} for results on scientific LLMs.

\subsection{Factual Accuracy and Model Confidence}
\label{sec:factual_correct_and_model_confidence_measurement_methods}
\paragraph{Factual Accuracy}
For the Claim Judgment task, we map the model’s predictions $\hat{y}(c, t)$ and $\hat{y}(c)$ to the set $\{\textsc{support}, \textsc{refute}\}$ using manually identified answer patterns, and compare them against the ground-truth labels $y(c, t)$ and $y(c)$, respectively.
For the Claim Generation task, we assess the factual accuracy of the generated claim $\hat{c}$ by inviting \textsc{GPT-4o} to determine whether $y(\hat{c}, t) = \textsc{support}$, based on the abstract of the corresponding paper.

\paragraph{Model Confidence}
Given the absence of a validation set, we estimate confidence levels using three rule-based measurement methods and finalize the decision through majority voting. 
Rather than prescribing a single approach, our framework is designed to be flexible: we leave the choice of confidence estimation methods to users or practitioners, allowing them to select strategies that best align with their specific models and evaluation settings.

\begin{itemize}
\item \textbf{More Information} Following existing prompt-based solutions~\citep{feng2023knowledge, feng2024don}, we append a prompt asking whether more information is needed to answer a given question: \emph{``Do you need more information to answer this question? (Yes or No)''}. Indicating the need for more information suggests a lack of confidence.
\item \textbf{Consistency} 
We paraphrase the question three times, sample one response per version, and classify the model as confident only if all three responses give the same binary answer.
\item \textbf{Linguistic Confidence} Given only the model's response, we prompt \textsc{GPT-4o}~\citep{openai2024gpt4o} with the following question: \emph{``Do you think the model is confident about its answer? (Yes or No)''}, aiming to capture implicit linguistic markers of confidence, such as assertive phrasing, authoritative tone, and decisive language in the response. We also conduct a human evaluation of linguistic confidence. The confidence classification consistency between three human evaluators and \textsc{GPT-4o} is 75.9\%, thereby validating the effectiveness of \textsc{GPT-4o} as a judge in this task.
\end{itemize}
All three methods are used as confidence measurement for the judgment responses, while \textbf{More Information} is used for generation responses. 

\subsection{Knowledge Update Methods}
\label{sec:knowledge_upadte_methods}
We experiment with five knowledge update methods that update new scientific knowledge (i.e., $\mathcal{P}_\text{new}$) at either the training stage, the inference stage, or both. $\mathcal{P}_\text{new}$ is split into training and test sets, and only the test set will be evaluated. Following previous work on scientific domains~\citep{wadden2020fact, newman2024arxivdigestables}, we use abstracts of papers in $\mathcal{P}_\text{new}$ instead of full papers, as they are easier to fit within the model's context window and align with practical constraints related to data availability and copyright restrictions.

\paragraph{Training.}
Through training, we update the model parameters by minimizing loss defined by different training objectives. Only LoRA adapters~\citep{hu2021lora} are trained for all training baselines, with the training duration set to 1 epoch for autoregressive training and 4 epochs for SFT.

\begin{itemize} [label=,leftmargin=0.5cm]
\item{\bf Continual Pre-training (\textsc{cnt pretrain}).} $\mathcal{P}_\text{new}^{\text{test}}$ is introduced through autoregressive training~\citep{gururangan2020don}, minimizing the standard next-token prediction loss: $-\frac{1}{|d|} \sum_{t} \log p_{\theta}(d_t | d_{<t})$. 

\item{\bf Standard Instruction-tuning (\textsc{inst tune}).} The model is first trained autoregressively on both $\mathcal{P}_{\text{new}}^\text{train}$ and $\mathcal{P}_{\text{new}}^\text{test}$, and then fine-tuned~\citep{wei2021finetuned} on training QA by minimizing the answer prediction loss given the question: $-\frac{1}{|a|} \sum_{t} \log p_{\theta}(a_t | q, a_{<t})$.

\item{\bf Pre-instruction-tuning (\textsc{pre inst tune}).} \citet{jiang2024instruction} exposes LLMs to QA pairs before continued pre-training on documents. Specifically, the model is instruction-tuned on training QA along with $\mathcal{P}_{\text{new}}^\text{train}$ prior to autoregressively trained on $\mathcal{P}_{\text{new}}^\text{test}$.
\end{itemize}

\paragraph{Inference (\textsc{infer}). }
The success of in-context learning~\citep{brown2020language} highlights the potential for introducing new knowledge at inference time, offering a more cost-efficient approach. Many existing knowledge augmentation methods, including RAG~\citep{shi2023replug}, search engines~\citep{Press2022MeasuringAN}, and multi-LLM collaborations~\citep{feng2023knowledge, feng2024multiLLM}, leverage this strategy to provide additional information. In our setting, we add corresponding $p_\text{new}$ in $\mathcal{P}_{\text{new}}^\text{test}$ to the prompt text.

\paragraph{Training + Inference (\textsc{inst tune} + \textsc{infer}).}
Following \citet{tang2024evowiki}, we also explore whether combining training and inference-time methods can yield improved performance. Specifically, we integrate standard instruction-tuning with the inference-time approach.

\subsection{Results}
\label{sec:results}

\begin{table*}[t]
    \centering
    \setlength{\tabcolsep}{1.5pt}
    \resizebox{\textwidth}{!}{
    \begin{tabular}{lccc|ccc|cc||ccc|ccc|cc}
         \toprule[1.5pt]
         \multirow{3}{*}{\textbf{Method}} & \multicolumn{8}{c}{\textbf{Claim Judgment Task}} & \multicolumn{8}{c}{\textbf{Claim Generation Task}} \\
          &
          \textcolor{palevioletred}{\textbf{Pres}} &
          \textcolor{palevioletred}{Dist} &
          \textcolor{palevioletred}{Loss} &
          \textcolor{lightblue}{\textbf{Acqu}} &
          \textcolor{lightblue}{Dist} & 
          \textcolor{lightblue}{Loss} & 
          \textcolor{goldenrod}{\textbf{Proj}}& 
          \textcolor{goldenrod}{Loss}& 
          \textcolor{palevioletred}{\textbf{Pres}} & 
          \textcolor{palevioletred}{Dist} &
          \textcolor{palevioletred}{Loss} &
          \textcolor{lightblue}{\textbf{Acqu}} &
          \textcolor{lightblue}{Dist} & 
          \textcolor{lightblue}{Loss} & 
          \textcolor{goldenrod}{\textbf{Proj}}& 
          \textcolor{goldenrod}{Loss} \\ \midrule[0.75pt]
         \multicolumn{17}{c}{\textbf{\textsc{LLaMA3.1-8B-Instruct}} }\\ \midrule[0.75pt]
         \textcolor{black}{\textsc{cnt pretrain}} & \underline{85.0} & \underline{5.5}& \underline{9.5}& 37.3 & 29.9& 32.8& 34.5& 48.3& 53.3&30.0& 16.7&53.1 & 42.0& \textbf{5.0}& 11.8&70.6\\
         \textcolor{black}{\textsc{inst tune}} & \textbf{86.3} & \textbf{4.1}& 9.6& 38.9 & \underline{28.3}& 32.8 & 24.1& 41.3&\textbf{72.2}& \textbf{17.8}&\textbf{10.0} & 56.1& 38.2 & \underline{5.7} &\textbf{29.4}&64.7\\
         \textcolor{black}{\textsc{pre inst tune}} & 59.0 & 38.3& \textbf{2.7}& \textbf{64.2} & \textbf{26.8}& \underline{9.0} & \underline{44.9}& 48.2& \underline{63.3}& \underline{23.3}& \underline{13.3}& 56.1&37.4& 6.5&11.8& 64.7\\
         \textcolor{black}{\textsc{infer}} & 68.6 & 17.8& 13.6& \underline{43.2} & 50.8& \textbf{6.0} & \textbf{48.3}& \underline{13.7}& 14.4& 62.2& 23.3& \textbf{84.4}&\textbf{8.4}& 7.3&\underline{23.5}&\textbf{5.9}\\
         \textcolor{black}{\textsc{inst tune}} + \textcolor{black}{\textsc{infer}} & 69.9 & 19.2& 10.9& 41.8 & 43.3& 15.0 & \underline{44.9}& \textbf{6.8}& 12.2& 58.9&28.9&\underline{76.0} &\underline{11.5}&12.6 & 17.6&\underline{17.6}\\
         \midrule[0.75pt]
         \multicolumn{17}{c}{\textsc{\textbf{OLMo2-32B-Instruct}}}\\ \midrule[0.75pt]
         \textcolor{black}{\textsc{cnt pretrain}} & 89.4 &\textbf{0.0}& 10.6& 18.7&40.7& \underline{40.7}& 16.6& 63.8& 82.5&17.5& \textbf{0.0}& 68.3&31.7&\textbf{0.0} &13.1&71.5\\
         \textcolor{black}{\textsc{inst tune}} & 89.5&\underline{0.9}& 9.6& 20.3&35.6& 44.2& 13.8& 68.5&\textbf{85.8}& \textbf{14.2}&\textbf{0.0}&67.7&32.3 &\textbf{0.0} &18.9&71.3\\
         \textcolor{black}{\textsc{pre inst tune}} & 89.4&\underline{0.9}& 9.7& 17.0&39.9& 43.2& 17.6& 65.7&\underline{84.2}&\underline{15.8} &\textbf{0.0}&68.3&31.7 &\textbf{0.0}&18.6&63.9\\
         \textcolor{black}{\textsc{infer}} & \textbf{99.1}&\underline{0.9}& \textbf{0.0}& \textbf{57.7}&\textbf{3.3}& \textbf{39.0}& \textbf{35.3}& \textbf{15.6} &42.9&55.8 &\underline{1.3}&\underline{79.3}& \underline{9.9}& \underline{10.8}&\textbf{37.6}&\textbf{13.7}\\
         \textcolor{black}{\textsc{inst tune}} + \textcolor{black}{\textsc{infer}} & \underline{96.1}&\underline{0.9}& \underline{2.9}& \underline{46.6}&\underline{6.8}& 46.7& \underline{33.3}& \underline{18.7}& 41.7&57.1&\underline{1.3}& \textbf{80.5}&\textbf{8.7} &\underline{10.8}&\underline{30.4}&\underline{26.4}\\
         \bottomrule[1.5pt]
    \end{tabular}
    }
    \caption{\textcolor{palevioletred}{Preservation}, \textcolor{lightblue}{Acquisition}, and \textcolor{goldenrod}{Projection} performance of knowledge update methods in the domain of \emph{Computer Science}. Best results in \textbf{bold} and second best in \underline{underline}. 
    Since we take model confidence into account (Section~\ref{sec:framework}), the random baseline is below 50.
    Higher values of preservation, acquisition, and projection are better, while lower values of \texttt{distortion} and \texttt{loss} are preferred.
    All methods fail to meet objectives collectively.}
    \label{tab:main}
\end{table*}

\paragraph{No knowledge update method can simultaneously achieve all three objectives.}
As shown in Table~\ref{tab:main}, the best-performing knowledge update methods, averaged across tasks and models, preserve only 85.9\% of existing knowledge, acquire 71.7\% of new knowledge, and project 37.7\% (or more) of future knowledge. However, we fail to find a method that can achieve all three objectives collectively. We additionally provide results on factual accuracy alone, without taking confidence into account, in Appendix~\ref{sec:noconf} for reference. Overall, standard instruction-tuning and inference methods remain as the strongest methods across all five.
Enabling models to project future knowledge presents a new challenge for knowledge update. As LLMs become increasingly integrated into scientific workflows such as hypothesis and idea generation, it becomes critical to develop update methods that not only integrate new claims but also enable models to anticipate and reason about future advancements.

\paragraph{Inference-based methods work for larger models, whereas smaller models require training to achieve comparable performance.}
For instance, in the claim judgment task, OLMo2-32B achieves inference-time performance that is 10.9\% higher than training-based methods in Knowledge Preservation on average, whereas the inference-time method on LLaMA-8B performs 10.7\% worse than training-based methods.
This discrepancy arises in part from the larger models' capacity to effectively filter out irrelevant or noisy contextual information during inference. With their greater representational power and more robust internal attention mechanisms, larger models are less sensitive to distractions in the prompt or context~\citep{brown2020language, askell2021general, min2022rethinking}, allowing them to incorporate new knowledge with minimal degradation of prior understanding.
From a computational perspective, inference-based update is significantly more cost-effective than training~\citep{houlsby2019parameter, rebuffi2017learning, lester2021power}, especially when updates are frequent. Notably, combining inference with additional training does not lead to performance improvements over inference alone, suggesting diminishing returns from training once a model has sufficient capacity to leverage inference-based strategies effectively. For smaller models, however, training remains a necessary step to compensate for their limited ability to generalize and disambiguate new knowledge in context. More in Appendix~\ref{sec:other-models}.

\paragraph{In the more challenging task, \texttt{distortion} is significantly greater than \texttt{loss}.}
Specifically, in the Generation task, the amount of \texttt{distortion} is, on average, three times higher than \texttt{loss} in both Preservation and Acquisition.
As we discussed in Section~\ref{sec:evaluation_of_knowledge_update_methods}, \texttt{distortion} is considered more severe than \texttt{loss} in both scenarios. 
This observation suggests a significant challenge for knowledge update methods, as they may introduce errors even when they should remain cautious. To address this issue, future developments in knowledge update methods could incorporate an abstention mechanism, avoiding updating models' representations if they lack confidence or certainty about the new content. Such a mechanism would allow models to opt out of updates when faced with ambiguous or uncertain knowledge, helping to preserve accuracy and reduce the propagation of errors.


\section{Cross-domain Analysis}
\label{sec:domains}

\begin{figure*}[t]
    \centering
    \vspace*{-1mm}
    \includegraphics[width=1.0\textwidth]{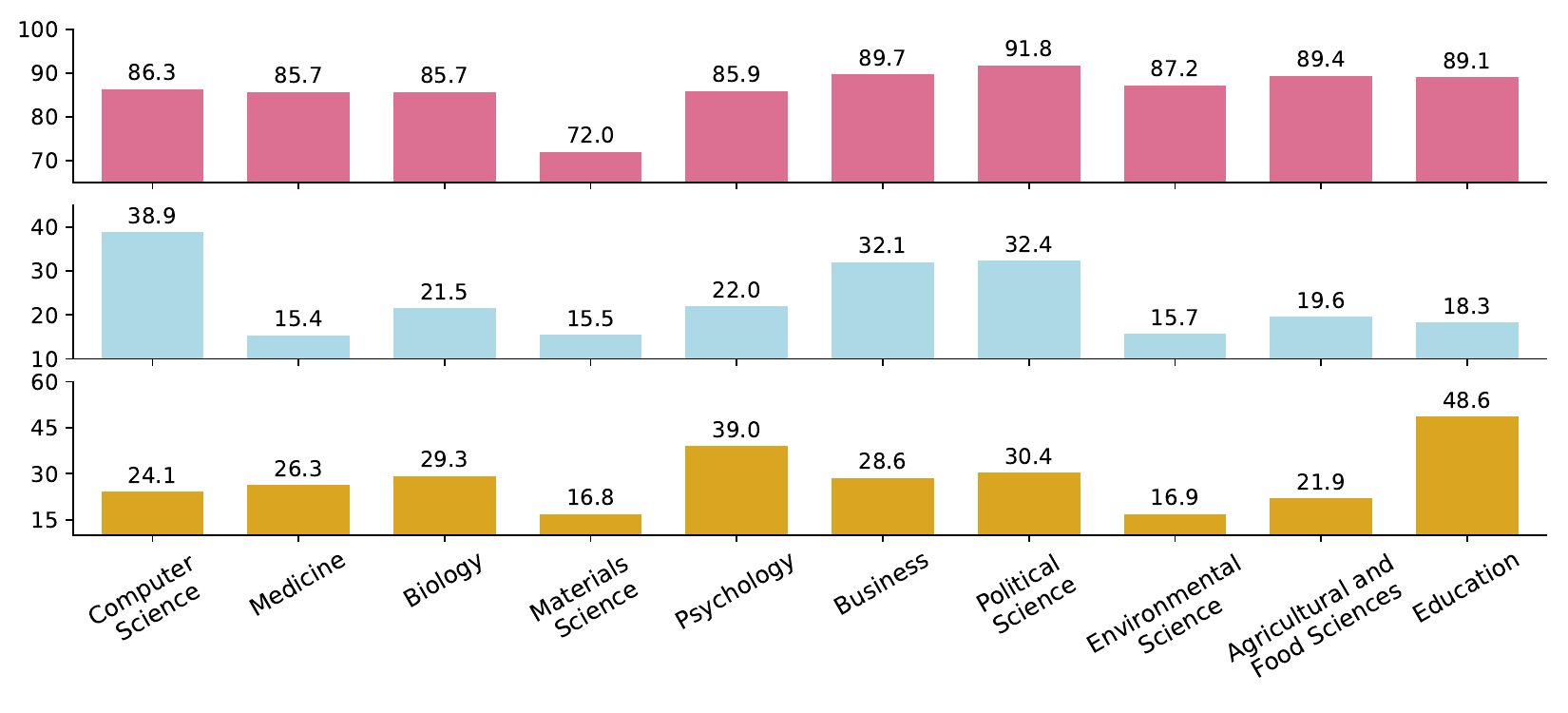}
    \vspace*{-11mm}
    \caption{
    \textcolor{palevioletred}{Preservation}, \textcolor{lightblue}{Acquisition}, and \textcolor{goldenrod}{Projection} performance of Standard Instruction-tuning on \textsc{LLaMA3.1-8B} in the claim judgment task. 
    The performance in \emph{Materials Science} and \emph{Environmental Science} is poor across all three objectives, whereas \emph{Political Science} and \emph{Education} show relatively strong performance in all three.}
    \label{fig:domain}
\end{figure*}

\begin{figure*}[t]
    \centering
    \includegraphics[width=1.0\textwidth]{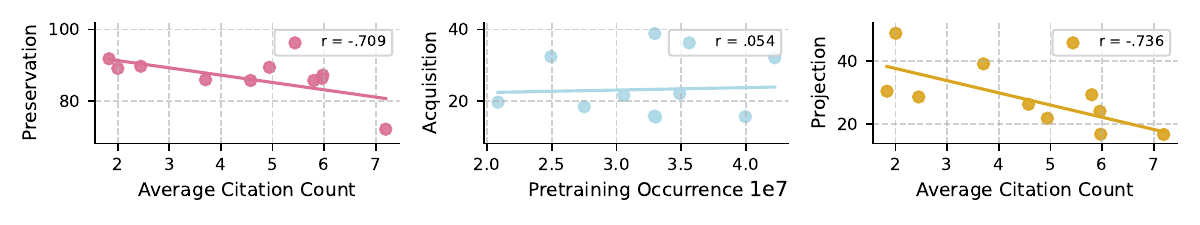}
    \vspace*{-8mm}
    \caption{The correlation between \textcolor{palevioletred}{Preservation}, \textcolor{lightblue}{Acquisition}, \textcolor{goldenrod}{Projection} and average citation count, as well as pretraining occurrence. The closer the points are to the best‑fit line, the stronger the correlation.}
    \label{fig:domain-factors}
\end{figure*}

In this section, we further break down performance by scientific domain. As shown in Figure~\ref{fig:domain}, performance varies significantly across domains. While over 90\% of scientific knowledge in \emph{Political Science} is preserved, only 72\% of \emph{Materials Science} knowledge can be retained. Similarly, while 48.6\% of scientific knowledge in \emph{Education} can be projected, this drops to just 16.8\% in \emph{Materials Science}. We also notice that these three capabilities appear to be correlated. Performance in certain domains tends to be consistently poor, for example, \emph{Materials Science} and \emph{Environmental Science}, whereas domains such as \emph{Political Science} exhibit relatively strong performance across all three objectives. 

We hypothesize that the domain performance might be influenced by two key factors:

First, the nature of the domain, specifically the stability or volatility of domain knowledge. 
Knowledge preservation and projection may be more challenging in domains with higher volatility (such as \emph{Materials Science}, with frequent breakthroughs or shifting paradigms that rapidly change the state of knowledge) compared to those with greater stability (such as \emph{Political Science}, with long-established theories and principles that evolve slowly over time). 
To test this hypothesis, we randomly retrieve 1,000 papers published during the same time period in each domain, and calculate the average citation count for these papers (Appendix~\ref{sec:cross-domain-analysis}), under the assumption that higher average citation counts reflect higher knowledge volatility.

Second, the availability of domain knowledge in the pretraining corpus. We posit that if domain knowledge appears frequently or widely in the pretraining corpus, knowledge acquisition might be easier. To assess this possibility, we collect the 100 tokens that appear least frequently in the abstracts of these 1,000 papers in each domain, as they tend to be specialized tokens unique to each domain. We then use Infini-gram~\citep{Liu2024InfiniGram} to count the occurrence of these tokens in the pretraining corpus Dolma-v1.7~\citep{dolma}, and use the average occurrence of domain-specific tokens as a proxy for the availability of domain knowledge in the pretraining data (Appendix~\ref{sec:cross-domain-analysis}). 

As shown in Figure~\ref{fig:domain-factors}, our analysis reveals a strong relationship between the ability to preserve and project scientific knowledge and the dynamics of the domain. Specifically, the Pearson correlation coefficients~\citep{pearson1895} between average citation count and Knowledge Preservation and Projection are -0.709 and -0.736 respectively, both indicating a significant relationship. Highly dynamic domains with frequent updates may lead to more knowledge conflicts~\citep{wang2023resolving}, making preservation and projection particularly challenging.
In contrast, the correlation between pretraining availability and Knowledge Acquisition is relatively weak, indicating that pretraining alone may have a limited impact on how well models adapt to evolving scientific information. 

\section{Related Work}
\label{sec:related_work}
\paragraph{LLMs for Scientific Advancements}
Recent research has demonstrated the significant potential of LLMs in driving scientific advancements across various domains, revolutionizing the way researchers approach complex problems and innovate in their respective fields~\citep{liang2024mapping, luo2025llm4sr, hsu2024chime, qi2023large, jansen2025discoveryworld, ahn2024transformative, si2024can, starace2025paperbench}.
Researchers utilize off-the-shelf LLMs~\citep{ai4science2023impactlargelanguagemodels}, domain-specific scientific LLMs~\citep{zhang2024comprehensive}, or LLMs augmented with external resources~\citep{asai2024openscholar} to assist in scientific research.
Studies show that current LLMs can be useful across various stages of the research cycle~\citep{luo2025llm4sr}, including literature review~\citep{hsu2024chime, agarwal2024litllm}, hypothesis proposing~\citep{qi2023large}, idea generation~\citep{si2024can}, and experiment planning and implementation~\citep{jansen2025discoveryworld, huang2023benchmarking}. 
However, to the best of our knowledge, we are the first to examine whether LLMs can stay up to date and continuously contribute to the advancement of these fields.

\paragraph{Evaluation of Knowledge Updates in LLMs}
Our evaluation of scientific knowledge updates differs from existing work on evaluation of knowledge updates in LLMs~\citep{li2025memorization, sun2025new, wang2024lekube} in three key aspects. 
First, most prior work primarily regards the effective incorporation of new information as the only objective~\citep{jang2021towards, ovadia2023fine, jiang2024instruction, zhang2024self, tang2024evowiki, yin2024history, zhao2024set, jang2022temporalwiki}, with few also considering the preservation of old knowledge~\citep{jiang2024instruction, zhang2024self}. However, they rely on generic benchmarks such as Natural Questions~\citep{kwiatkowski2019natural} and CommonsenseQA~\citep{talmor2018commonsenseqa}, which evaluate the retention of general world knowledge rather than the preservation of knowledge related to the newly updated information. And we further introduce a new evaluation dimension, evaluating the utility of knowledge updates for reasoning, hypothesis generation~\citep{qi2023large}, and the creation of novel ideas~\citep{si2024can}, which are the key future applications of AI in science.
Second, existing approaches heavily rely on Wikipedia as the data source and assess knowledge at the factoid level (e.g., names, locations)~\citep{jang2021towards, ovadia2023fine, jiang2024instruction, zhang2024self, tang2024evowiki, yin2024history, zhao2024set, jang2022temporalwiki}, whereas we extend evaluation to natural language representations, which better capture the core insights and implications of research beyond isolated numerical values as well as the complexity of real-world knowledge.
Third, prior work on knowledge alignment primarily focuses on temporal alignment~\citep{zhao2024set, yin2024history, jang2022temporalwiki}, aiming to align knowledge to specific timestamps, such as associating a president with a particular year, while our goal is to align scientific claims with scientific literature. 

We distinguish our evaluation of knowledge preservation from Catastrophic Forgetting (CF) in Continual Learning (CL)~\citep{kirkpatrick2017overcoming, jang2021towards, clemente2025stubbornness}, as our setting involves multiple training stages and methods. Another relevant line of work is knowledge editing~\citep{meng2022mass, wang2024knowledge, meng2022locating, zhang2024comprehensive, liu2024codeupdatearena, wang2024wise, jiang2024learning, liu2025mitigating, he2025knowledge}, which aims to replace incorrect existing knowledge, whereas our goal is to integrate new knowledge without altering the model’s understanding of previously learned scientific literature.

\section{Conclusion}
\label{sec:conclusion}
In this work, we investigate scientific knowledge updates of LLMs and propose that an effective and reliable update method should be able to preserve existing scientific knowledge, acquire new scientific knowledge, and project future scientific knowledge, which are crucial for the continual use of LLMs in evolving scientific fields. 
To this end, we introduce an evaluation framework \methodname with rich datasets of scientific papers across domains, new tasks and evaluation of scientific knowledge, and new metrics for evaluating knowledge update methods.
With comprehensive experiments on frontier general-purpose and science-focused LLMs, we find that achieving these objectives remains an open research challenge, underscoring the need for further exploration in this direction.

\section*{Limitations}
\label{sec:limitations}

\paragraph{Real scientific advancement is far more complex}
In this work, we model scientific advancement as a linear timeline spanning existing, new, and future developments. However, genuine scientific progress is considerably more complex in reality. New advancements often emerge from the convergence of multiple research trajectories across diverse domains. Future work should aim to capture this multidimensional nature of scientific progress.

\paragraph{Beyond scientific claims}
While this work focuses on scientific claims as the fundamental unit of analysis for evaluating scientific knowledge and scientific knowledge updates, we recognize that scientific knowledge operates at multiple meaningful levels of granularity. Other critical dimensions worthy of investigation include the paper-level and researcher-level, which could be potential directions for future research.

\paragraph{Contradictory claims}
When evaluating future scientific knowledge using claim classification tasks, we acknowledge that, theoretically, there is a chance that some claims may contradict past findings. However, empirical evidence suggests such occurrences are rare. Moreover, our claims are sufficiently detailed and comprehensive, making it unlikely that identical or highly similar claims have appeared in prior literature.

\paragraph{Disentangling knowledge from instruction-following and reasoning capabilities}
Separating the ``knowledge'' of LLMs from their instruction-following and reasoning abilities is challenging, particularly if we define ``knowledge'' as the content they generate. Prior work has attempted to assess LLMs' knowledge using cloze-style tasks~\citep{jang2021towards} at inference time, which rely more on raw knowledge and less on instruction-following ability. However, such formats are limited to evaluating factoid knowledge. In this work, we define scientific knowledge as scientific claims and propose claim judgment and generation tasks to evaluate it. While these tasks are effective for assessment and analysis, we acknowledge that model performance on them still depends, to some extent, on instruction-following and reasoning capabilities.

\section*{Acknowledgments}
This research was developed with funding from the Defense Advanced Research Projects Agency's (DARPA) SciFy program (Agreement No. HR00112520300), and NSF IIS-2044660. The views expressed are those of the authors and do not reflect the official policy or position of the Department of Defense or the U.S. Government. This work was also supported in part by Azure credits from Microsoft Accelerate Foundation Models Research.

\section*{Ethics Statement}
\label{sec:ethics_statement}
We envision certain potential ethical risks of \methodname. For example, when evaluating ``future'' scientific claims, the framework risks creating ethical dilemmas regarding premature validation of unproven hypotheses. This becomes particularly problematic when assessing claims in sensitive domains (e.g., climate science or medical research) where premature endorsement could influence policy or clinical decisions.

However, \methodname also provides significant ethical benefits by introducing systematic transparency to scientific knowledge assessment. \methodname can help surface meritorious but underrecognized research directions, and these features may ultimately promote more equitable and evidence-based scientific progress when implemented with appropriate ethical safeguards.

\bibliography{colm2026_conference}
\bibliographystyle{colm2026_conference}

\clearpage
\appendix

\section{Dataset Details}
\label{sec:dataset_details}

\subsection{Date Cutoffs}
Table~\ref{tab:date-cutoffs} presents the specific date cutoffs used to construct the dataset for all models in this study. A three-month buffer accounts for potential discrepancies between a paper's online availability and its official publication date, allowing a more accurate representation of respective knowledge.

\begin{table*}[h!]
\begin{center}
\setlength{\tabcolsep}{3pt}
\begin{tabular}{lcccc}
\toprule[1.5pt]
 \textbf{Model} & \textbf{Cutoff}   & \textbf{Prior Knowledge} & \textbf{New Knowledge} & \textbf{Future Knowledge}\\
\midrule[0.75pt]
\textsc{LLaMA3.1-8B} & Dec 2023 & 2022.10.1 - 2023.9.30 & 2024.3.1 - 2024.11.30
 & 2024.12.1 - 2025.2.1\\
 \textsc{OLMo2-7B} & Dec 2023 & 2022.10.1 - 2023.9.30 & 2024.3.1 - 2024.11.30
 & 2024.12.1 - 2025.2.1\\
\textsc{OLMo2-32B} & Dec 2023 & 2022.10.1 - 2023.9.30 & 2024.3.1 - 2024.11.30
 & 2024.12.1 - 2025.2.1\\
 \textsc{HoneyBee} & Oct 2023 & 2022.8.1 - 2023.7.31 & 2024.1.1 - 2024.11.30
 & 2024.12.1 - 2025.3.1\\
\bottomrule[1.5pt]
\end{tabular}
\end{center}
\caption{Date cutoffs used to distinguish prior, new, and future knowledge when constructing the dataset for different models.}
\label{tab:date-cutoffs}
\end{table*}

\subsection{Metadata}
Table~\ref{tab:metadata} shows the number of papers in each domain in our dataset.

\begin{table}[h!]
\centering
\setlength{\tabcolsep}{2pt}
\begin{tabular}{lc}
\toprule[1.5pt]
\textbf{Domain} & \textbf{Paper Count} \\
\midrule[0.75pt]
Computer Science & 835 \\
Medicine & 480 \\
Biology & 351 \\
Materials Science & 559 \\
Psychology & 491 \\
Business & 503 \\
Political Science & 409 \\
Environmental Science & 455 \\
\makecell[l]{Agricultural and\\Food Sciences} & 533 \\
Education & 532 \\
\midrule[0.75pt]
SUM & 5148 \\
\bottomrule[1.5pt]
\end{tabular}
\caption{Paper Count in each domain after filtering out papers without citation information or abstracts. Originally, 1,000 papers were retrieved per domain.}
\label{tab:metadata}
\end{table}

\subsection{Synthetic Claims}
To generate synthetic claims for each paper, we prompt \textsc{GPT-4o} with the instructions detailed in Table~\ref{tab:prompts-synthetic-claims}.
We explored various alternative methods for synthetic claim generation, such as retrieving a relevant paper and using its \textsc{support} claim as the \textsc{refute} claim for the given paper. However, the method we ultimately adopted, despite its simplicity, yielded the best results. Additionally, we control the granularity of claims by constraining their length to approximately 15 words, ensuring that they are neither overly simplistic nor excessively verbose (e.g., the entire abstract).

\begin{table*}
\begin{tcolorbox}[title=Prompt: \textsc{support} Claim Generation]
{\bf System Prompt}

You are an expert scientific research assistant.

\tcblower

{\bf User Prompt} 

Please identify and extract the main scientific claim that is uniquely supported by the given paper. A scientific claim is a atomic verifiable statements expressing a finding about one aspect of a scientific entity or process, which can be verified against a single source.

\end{tcolorbox}

\begin{tcolorbox}[title=Prompt: \textsc{refute} Claim Generation]
{\bf System Prompt}

You are an expert scientific research assistant.

\tcblower

{\bf User Prompt} 

Please identify and extract a scientific claim that is relevant but not supported by the given paper. A scientific claim is a atomic verifiable statements expressing a finding about one aspect of a scientific entity or process, which can be verified against a single source.

\end{tcolorbox}
\caption{Prompt templates for synthetic claim generation.}
\label{tab:prompts-synthetic-claims}
\end{table*}

We further conduct an expert evaluation of our synthetic claims.
We invite 20 PhD students across ten domains and assign each student 50 papers within their respective area of expertise. 
For each paper, we provide the title, abstract, and two synthetic claims, and they are instructed to classify each claim into one of the following categories:  

\begin{itemize}
    \item \textbf{Uniquely Supported} – The claim can only be verified by the given paper.
    \item \textbf{Broadly Supported} – The claim is supported by the given paper but is likely validated by other sources as well.
    \item \textbf{Not Supported} – The claim is not supported by the given paper.
\end{itemize}

The results are presented in Table~\ref{tab:synthetic_claims_expert_evaluation}, showing that, on average, 90.2\% of claims strictly adhere to the rule, while 98.7\% broadly meet our expectations. While the results demonstrate the effectiveness of our synthetic claims, there is still room for improvement, so we try to collect author-annotated claims in Section~\ref{sec:author_annotated_claims}.


\begin{table*}
\begin{center}
\small
\begin{tabular}{llccc}
\toprule[1.5pt]
\textbf{Domain} & \textbf{Label} 
& \textbf{\makecell{Uniquely\\Supported}} 
& \textbf{\makecell{Broadly \\ Supported}} 
& \textbf{\makecell{Not \\ Supported}} \\
\midrule[0.75pt]

\multirow{2}{*}{Computer Science} 
 & \textsc{support} & 92\% & 8\% & 0\% \\
 & \textsc{refute}  & 0\%  & 10\% & 90\% \\

\midrule[0.75pt]
\multirow{2}{*}{Medicine} 
 & \textsc{support} & 86\% & 12\% & 2\% \\
 & \textsc{refute}  & 2\%  & 8\% & 90\% \\

\midrule[0.75pt]
\multirow{2}{*}{Biology}          
 & \textsc{support} & 86\% & 14\% & 0\% \\
 & \textsc{refute}  & 0\%  & 12\% & 88\% \\

\midrule[0.75pt]
\multirow{2}{*}{Materials Science} 
 & \textsc{support} & 88\% & 8\% & 4\% \\
 & \textsc{refute}  & 2\%  & 14\% & 84\% \\

\midrule[0.75pt]
\multirow{2}{*}{Psychology} 
 & \textsc{support} & 98\% & 2\% & 0\% \\
 & \textsc{refute}  & 4\%  & 10\% & 86\% \\

\midrule[0.75pt]
\multirow{2}{*}{Business} 
 & \textsc{support} & 94\% & 4\% & 2\% \\
 & \textsc{refute}  & 0\%  & 18\% & 82\% \\

\midrule[0.75pt]
\multirow{2}{*}{Political Science} 
 & \textsc{support} & 86\% & 10\% & 4\% \\
 & \textsc{refute}  & 2\%  & 10\% & 88\% \\

\midrule[0.75pt]
\multirow{2}{*}{Environmental Science} 
 & \textsc{support} & 98\% & 2\% & 0\% \\
 & \textsc{refute}  & 0\%  & 2\% & 98\% \\

\midrule[0.75pt]
\multirow{2}{*}{\makecell[l]{Agricultural and\\Food Sciences}}
 & \textsc{support} & 94\% & 4\% & 2\% \\
 & \textsc{refute}  & 0\%  & 6\% & 94\% \\

\midrule[0.75pt]
\multirow{2}{*}{Education} 
 & \textsc{support} & 88\% & 10\% & 2\% \\
 & \textsc{refute}  & 0\%  & 6\% & 94\% \\

\bottomrule[1.5pt]
\end{tabular}
\end{center}
\caption{Expert evaluation results on synthetic claims, averaged across two experts per domain.}
\label{tab:synthetic_claims_expert_evaluation}
\end{table*}

\subsection{Author-annotated Claims}
\label{sec:author_annotated_claims}
\begin{figure*}[h!]
    \centering
    \includegraphics[width=1.0\textwidth]{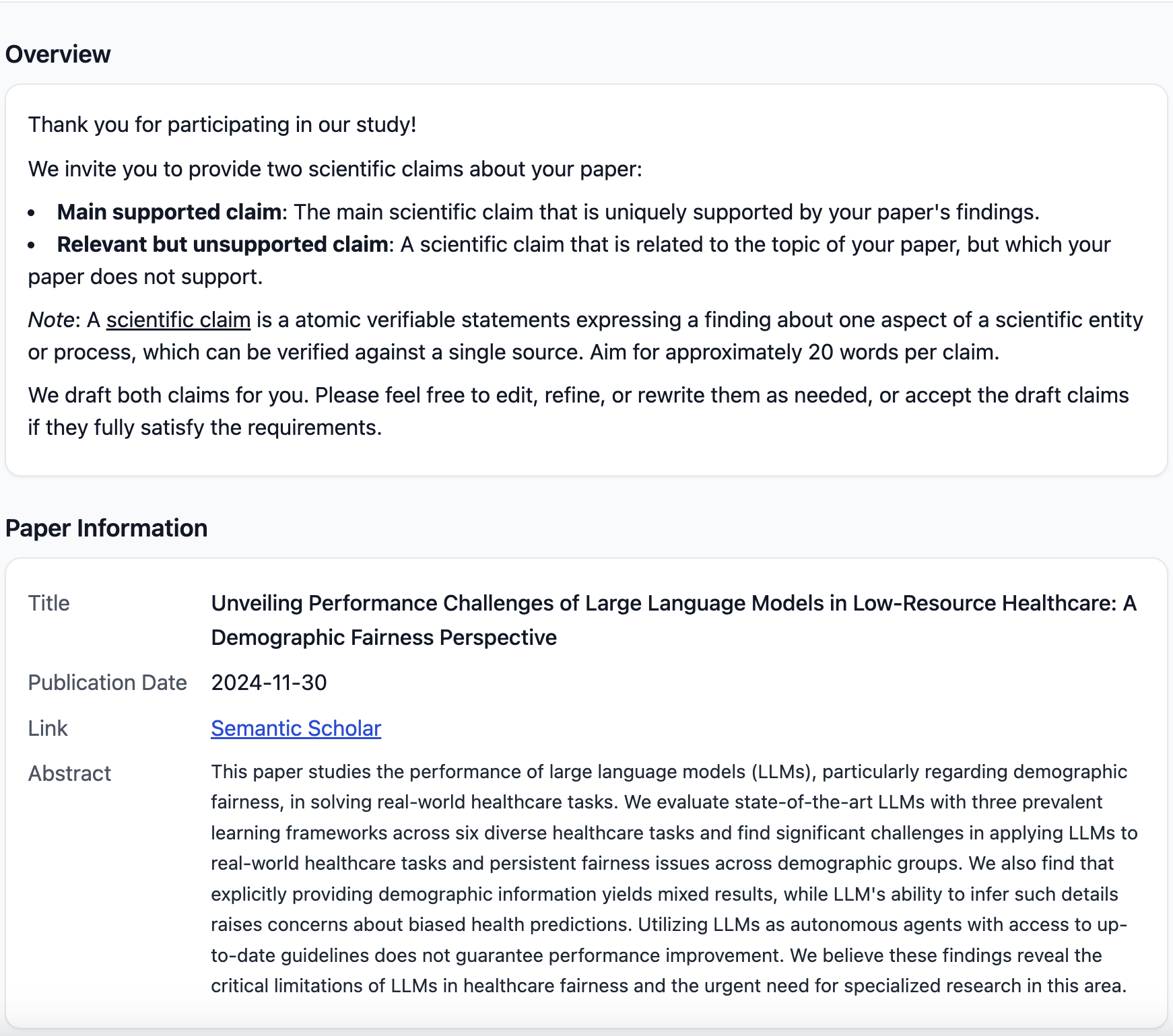}
    \caption{An example of the scientific claims author annotation interface.}
    \label{fig:annotation-interface}
\end{figure*}

\begin{figure*}[h!]
    \centering
    \includegraphics[width=1.0\textwidth]{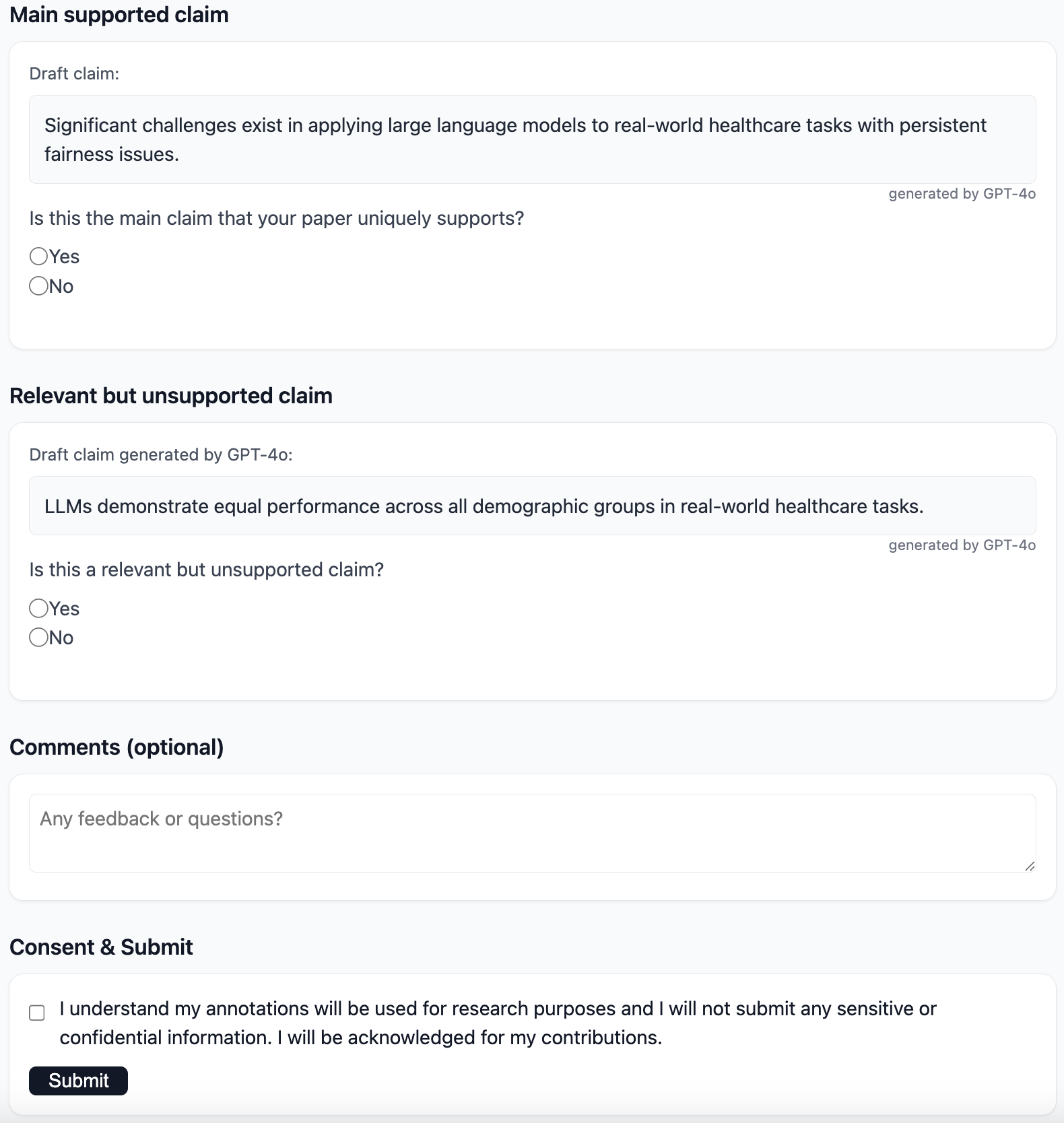}
    \caption{An example of the scientific claims author annotation interface (cont.). Authors can either accept the proposed claim or edit it in the edit boxes.}
    \label{fig:annotation-interface-cont}
\end{figure*}

As the original authors of research papers possess the most relevant expertise for claim annotation, we aim to collect author-annotated claims in the \textit{Computer Science} domain under a limited annotation budget. Specifically, we develop a scientific claim annotation interface (Figure~\ref{fig:annotation-interface} and Figure~\ref{fig:annotation-interface-cont}) and email the first authors of 500 papers from our \textit{Computer Science} dataset. Although some emails were undeliverable, we successfully received responses from 43 authors, resulting in 86 annotated scientific claims (43 labeled as \textsc{support} and 43 as \textsc{refute}).
Out of the 43 responses, 38 authors accepted the proposed \textsc{support} claims and 29 authors accepted the proposed \textsc{refute} claims. For the rejected cases, the authors edited and revised the claims.

Our evaluation using LLaMA-8B on this author-annotated dataset (Table~\ref{tab:author_annotated_claims}) reveals no statistically significant performance difference compared to synthetic claims. This finding empirically validates the effectiveness of our synthetic claim generation methodology, suggesting that the synthetic claims maintain comparable quality to human expert annotations while offering scalability advantages.

\begin{table*}[h!]
    \centering
    \setlength{\tabcolsep}{2pt}
    \vspace*{2mm}
    \resizebox{\textwidth}{!}{
    \begin{tabular}{lccc@{\hspace{6pt}}|ccc@{\hspace{6pt}}|cc} 
         \toprule[1.5pt]
          \textbf{Model} & 
          \textcolor{palevioletred}{\textbf{Preservation}} &
          \textcolor{palevioletred}{Distortion} &
          \textcolor{palevioletred}{Loss} &
          \textcolor{lightblue}{\textbf{Acquisition}} &
          \textcolor{lightblue}{Distortion} & 
          \textcolor{lightblue}{Loss} & 
          \textcolor{goldenrod}{\textbf{Projection}}& 
          \textcolor{goldenrod}{Loss}\\ 
          \midrule[0.75pt]
         \textsc{synthetic claims} & 86.3 & 4.1 & 9.6 & 38.9 & 28.3 & 32.8 & 24.1 & 41.3\\
         \textsc{author-annotated claims} & 89.1 & 3.2 & 7.7 & 34.2 & 27.5 & 38.3 & 22.6 & 37.2\\
         \bottomrule[1.5pt]
    \end{tabular}
    }
    \caption{We evaluate Standard Instruction-tuning on LLaMA-8B using our claim judgment task with synthetic and author-annotated claims in \emph{Computer Science} respectively. The results demonstrate no statistically significant difference between model performance on synthetic versus author-annotated claims, which validates the effectiveness of our synthetic claim generation approach.}
    \label{tab:author_annotated_claims}
\end{table*}

\section{Claim Judgment and Generation Tasks}
We present the prompts used for the claim judgment and generation tasks in Table~\ref{tab:prompts-claim-judgment-task} and Table~\ref{tab:prompts-claim-generation-task}.

\begin{table*}[h!]
\begin{tcolorbox}[title=Prompt: Claim Judgment Task - Claim Verification (Prior and New Knowledge)]
{\bf System Prompt}

You are an AI research assistant designed to provide accurate, evidence-based responses.  

\tcblower

{\bf User Prompt} 

claim: \{claim\}

Can every detail in the given claim be substantiated by the paper \{title\}?

\end{tcolorbox}

\begin{tcolorbox}[title=Prompt: Claim Judgment Task - Claim Classification (Future Knowledge)]
{\bf System Prompt}

You are an AI research assistant designed to provide accurate, evidence-based responses.  

\tcblower

{\bf User Prompt} 

claim: \{claim\}

Is the claim correct?

\end{tcolorbox}
\vspace*{-2mm}
\caption{Prompt templates for Claim Judgment Task.}
\label{tab:prompts-claim-judgment-task}
\end{table*}

\begin{table*}[h!]
\begin{tcolorbox}[title=Prompt: Claim Generation Task - Prior and New Knowledge]
{\bf System Prompt}

You are an AI research assistant designed to provide accurate, evidence-based responses.  

\tcblower

{\bf User Prompt} 

State the main scientific claim made in the paper \{title\}. A scientific claim is a atomic verifiable statements expressing a finding about one aspect of a scientific entity or process, which can be verified against a single source.

\end{tcolorbox}

\begin{tcolorbox}[title=Prompt: Claim Generation Task - Future Knowledge]
{\bf System Prompt}

You are an AI research assistant designed to provide accurate, evidence-based responses.  

\tcblower

{\bf User Prompt} 

State a scientific claim about \{subject\}. A scientific claim is a atomic verifiable statements expressing a finding about one aspect of a scientific entity or process, which can be verified against a single source.

\end{tcolorbox}
\vspace*{-2mm}
\caption{Prompt templates for Claim Generation Task.}
\label{tab:prompts-claim-generation-task}
\end{table*}

\section{Metrics}
\label{sec:metrics}

Table~\ref{tab:metrics} provides the detailed mathematical definitions of knowledge preservation, knowledge acquisition, and knowledge projection, as well as \texttt{distortion} and \texttt{loss}. 

\begin{table*}[h!]
\centering
\resizebox{\textwidth}{!}{
\renewcommand{\arraystretch}{2.3}
\setlength{\tabcolsep}{3pt}
\begin{tabular}{l}
\toprule[2.0pt] 
\textbf{Knowledge Preservation} 
$= \displaystyle \frac{\sum_{i} \mathbb{I}(g(LM_{f(\mathcal{P}_\text{new}^\text{test})}, p_\text{prior}^{i}) = \textit{correct} \mid g(LM, p_\text{prior}^{i}) = \textit{correct}, g(LM, p_\text{new}^{i}) = \textit{unknown}))}{\sum_{i} \mathbb{I}(g(LM, p_\text{prior}^{i}) = \textit{correct}, g(LM, p_\text{new}^{i}) = \textit{unknown})}$ \\[1.5ex]

\texttt{distortion} \text{in Preservation}
$= \displaystyle \frac{\sum_{i} \mathbb{I}(g(LM_{f(\mathcal{P}_\text{new}^\text{test})}, p_\text{prior}^{i}) = \textit{incorrect} \mid g(LM, p_\text{prior}^{i}) = \textit{correct}, g(LM, p_\text{new}^{i}) = \textit{unknown}))}{\sum_{i} \mathbb{I}(g(LM, p_\text{prior}^{i}) = \textit{correct}, g(LM, p_\text{new}^{i}) = \textit{unknown}))}$ \\[1.5ex]

\texttt{loss} \text{in Preservation}
$= \displaystyle \frac{\sum_{i} \mathbb{I}(g(LM_{f(\mathcal{P}_\text{new}^\text{test})}, p_\text{prior}^{i}) = \textit{unknown} \mid g(LM, p_\text{prior}^{i}) = \textit{correct}, g(LM, p_\text{new}^{i}) = \textit{unknown}))}{\sum_{i} \mathbb{I}(g(LM, p_\text{prior}^{i}) = \textit{correct}, g(LM, p_\text{new}^{i}) = \textit{unknown}))}$ \\[1.5ex]
\midrule[1.5pt]

\textbf{Knowledge Acquisition}
$= \displaystyle \frac{\sum_{i} \mathbb{I}(g(LM_{f(\mathcal{P}_\text{new}^\text{test})}, p_\text{new}^{i}) = \textit{correct} \mid g(LM, p_\text{new}^{i}) = \textit{unknown})}{\sum_{i} \mathbb{I}(g(LM, p_\text{new}^{i}) = \textit{unknown})}$ \\[1.5ex]

\texttt{distortion} \text{in Acquisition}
$= \displaystyle \frac{\sum_{i} \mathbb{I}(g(LM_{f(\mathcal{P}_\text{new}^\text{test})}, p_\text{new}^{i}) = \textit{incorrect} \mid g(LM, p_\text{new}^{i}) = \textit{unknown})}{\sum_{i} \mathbb{I}(g(LM, p_\text{new}^{i}) = \textit{unknown})}$ \\[1.5ex]

\texttt{loss} \text{in Acquisition}
$= \displaystyle \frac{\sum_{i} \mathbb{I}(g(LM_{f(\mathcal{P}_\text{new}^\text{test})}, p_\text{new}^{i}) = \textit{unknown} \mid g(LM, p_\text{new}^{i}) = \textit{unknown})}{\sum_{i} \mathbb{I}(g(LM, p_\text{new}^{i}) = \textit{unknown})}$ 
\\[1.5ex]
\midrule[1.5pt]

\textbf{Knowledge Projection}
$= \displaystyle \frac{\sum_{i} \mathbb{I}(g(LM_{f(\mathcal{P}_\text{new}^\text{test})}, p_\text{future}^{i}) = \textit{correct} \mid g(LM, p_\text{future}^{i}) = \textit{unknown}, g(LM, p_\text{new}^{i}) = \textit{unknown}))}{\sum_{i} \mathbb{I}(g(LM, p_\text{future}^{i}) = \textit{unknown}, g(LM, p_\text{new}^{i}) = \textit{unknown}))}$ \\[1.5ex]

\texttt{loss} \text{in Projection}
$= \displaystyle \frac{\sum_{i} \mathbb{I}(g(LM_{f(\mathcal{P}_\text{new}^\text{test})}, p_\text{future}^{i}) = \textit{unknown} \mid g(LM, p_\text{future}^{i}) = \textit{unknown}, g(LM, p_\text{new}^{i}) = \textit{unknown}))}{\sum_{i} \mathbb{I}(g(LM, p_\text{future}^{i}) = \textit{unknown}, g(LM, p_\text{new}^{i}) = \textit{unknown}))}$ \\[1.5ex]

\bottomrule[2pt]
\end{tabular}
}
\caption{Detailed formulations of evaluation metrics introduced in Section~\ref{sec:evaluation_of_knowledge_update_methods}. 
Note that $p_\text{prior}^i$ and $p_\text{future}^i$ are considered only if $p_\text{new}^i$ is \textit{unknown} to the model before knowledge updates, as otherwise no new scientific knowledge is introduced.}
\vspace{4mm}
\label{tab:metrics}
\end{table*}

\section{Experiment Details}
\label{sec:experiment_details}
We present the training configurations in Tables~\ref{tab:sft_config} and~\ref{tab:ar_config}. Experiments are performed on 4 A100 40GB GPUs.

\begin{table}[t]
\centering
\vspace*{2mm}
\label{tab:sft_config}
\begin{tabular}{lc}
\toprule
\textbf{Parameter} & \textbf{SFT} \\
\midrule
Learning Rate & $2 \times 10^{-4}$ \\
LR Scheduler & Cosine \\
Warmup Ratio & 0.1 \\
Per Device Train Batch Size & 1 \\
Gradient Accumulation Steps & 32 \\
Effective Batch Size & 32 \\
Precision & BF16 \\
\midrule
LoRA Rank ($r$) & 64 \\
LoRA Alpha ($\alpha$) & 16 \\
LoRA Dropout & 0.1 \\
LoRA Target Modules & \makecell{q\_proj, k\_proj, \\ v\_proj, o\_proj} \\
\midrule
Number of Epochs & 4 \\
\bottomrule
\end{tabular}
\caption{Training Configuration for SFT.}
\label{tab:sft_config}
\end{table}

\begin{table}[t]
\centering
\vspace*{2mm}
\label{tab:ar_config}
\begin{tabular}{lc}
\toprule
\textbf{Parameter} & \textbf{Value} \\
\midrule
Learning Rate & $2 \times 10^{-4}$ \\
LR Scheduler & Cosine \\
Warmup Ratio & 0.1 \\
Per Device Train Batch Size & 1 \\
Gradient Accumulation Steps & 32 \\
Effective Batch Size & 32 \\
Precision & BF16 \\
\midrule
LoRA Rank ($r$) & 64 \\
LoRA Alpha ($\alpha$) & 16 \\
LoRA Dropout & 0.1 \\
LoRA Target Modules & \makecell{q\_proj, k\_proj, \\ v\_proj, o\_proj} \\
\midrule
Number of Epochs & 1 \\
\bottomrule
\end{tabular}
\caption{Training Configuration for Autoregressive Training.}
\vspace*{4mm}
\label{tab:ar_config}
\end{table}

\section{Change in Factual Accuracy without Model Confidence}
\label{sec:noconf}
We also report the results of different knowledge update methods on the factual accuracy of prior knowledge in Table~\ref{tab:noconf}, without taking model confidence into account.

\begin{table*}[ht]
\centering
\small
\setlength{\tabcolsep}{3pt}
\begin{tabular}{lcccc|cccc}
\toprule[1.5pt]
 & \multicolumn{4}{c}{\textbf{Claim Judgment Task}} 
 & \multicolumn{4}{c}{\textbf{Claim Generation Task}} \\
\cmidrule(lr){2-5} \cmidrule(lr){6-9}
\textbf{Method} &
\shortstack{Accu.\\to\\Accu.} &
\shortstack{Accu.\\to\\Inaccu.} &
\shortstack{Inaccu.\\to\\Accu.} &
\shortstack{Inaccu.\\to\\Inaccu.} &
\shortstack{Accu.\\to\\Accu.} &
\shortstack{Accu.\\to\\Inaccu.} &
\shortstack{Inaccu.\\to\\Accu.} &
\shortstack{Inaccu.\\to\\Inaccu.} \\
\midrule[1.0pt]
\multicolumn{9}{c}{\textbf{\textsc{LLaMA3.1-8B-Instruct}}} \\
\midrule[1.0pt]
\textsc{cnt pretrain}        & 58.3 & 5.2  & 1.7  & 34.8 & 34.3 & 19.9 & 9.6  & 36.1 \\
\textsc{inst tune}          & 60.9 & 2.6  & 1.7  & 34.8 & 44.6 & 9.6  & 12.7 & 33.1 \\
\textsc{pre inst tune}      & 37.4 & 26.1 & 30.4 & 6.1  & 40.4 & 13.9 & 11.4 & 34.3 \\
\textsc{infer}              & 40.9 & 22.6 & 12.2 & 24.3 & 13.3 & 41.0 & 4.8  & 41.0 \\
\textsc{inst tune + infer}  & 46.1 & 17.4 & 1.7  & 34.8 & 12.0 & 42.2 & 2.4  & 43.4 \\
\midrule[1.0pt]
\multicolumn{9}{c}{\textbf{\textsc{OLMo2-32B-Instruct}}} \\
\midrule[1.0pt]
\textsc{cnt pretrain}        & 52.9 & 1.6 & 4.2 & 41.3 & 59.3 & 12.6 & 8.4 & 19.8 \\
\textsc{inst tune}          & 51.3 & 3.2 & 6.9 & 38.6 & 61.7 & 10.2 & 9.0 & 19.2 \\
\textsc{pre inst tune}      & 51.9 & 2.6 & 3.7 & 41.8 & 60.5 & 11.4 & 9.9 & 18.3 \\
\textsc{infer}              & 54.0 & 0.5 & 0.0 & 45.5 & 30.8 & 41.0 & 5.1 & 23.1 \\
\textsc{inst tune + infer}  & 53.4 & 1.1 & 0.5 & 45.0 & 29.9 & 41.9 & 6.9 & 21.3 \\
\bottomrule[1.5pt]
\end{tabular}
\caption{Changes in the factual accuracy of prior knowledge in the domain of \textit{computer science} under different models and settings.}
\label{tab:noconf}
\end{table*}

\section{Per-Method Breakdown of Confidence Estimators}
\label{app:confidence_breakdown}
 
Our confidence framework combines three complementary methods via majority voting: (1)~\textbf{More Information} prompting, (2)~\textbf{Consistency} across paraphrases, and (3)~\textbf{Linguistic Confidence} assessed by GPT-4o. We deliberately designed the framework to be flexible, so that practitioners can swap in alternative estimators depending on their model and setting as discussed in Section~\ref{sec:factual_correct_and_model_confidence_measurement_methods}.
 
In this section, we provide a per-method breakdown of results for LLaMA3.1-8B-Instruct on the Claim Judgment task under three representative knowledge update methods. Table~\ref{tab:confidence_breakdown} reports the full results.
 
\begin{table*}[t]
\centering
\vspace*{1mm}
\label{tab:confidence_breakdown}
\resizebox{\textwidth}{!}{%
\begin{tabular}{ll ccc ccc cc}
\toprule
& & \multicolumn{3}{c}{\textbf{Preservation}} & \multicolumn{3}{c}{\textbf{Acquisition}} & \multicolumn{2}{c}{\textbf{Projection}} \\
\cmidrule(lr){3-5} \cmidrule(lr){6-8} \cmidrule(lr){9-10}
\textbf{Method} & \textbf{Confidence Estimator} & \textcolor{palevioletred}{\textbf{Pres}} & \textcolor{palevioletred}{Dist} & \textcolor{palevioletred}{Loss} & \textcolor{lightblue}{\textbf{Acqu}} & \textcolor{lightblue}{Dist} & \textcolor{lightblue}{Loss} & \textcolor{goldenrod}{\textbf{Proj}} & \textcolor{goldenrod}{Loss} \\
\midrule
\multirow{4}{*}{\textsc{cnt pretrain}}
 & \textit{Majority vote} & 85.0 & 5.5 & 9.5 & 37.3 & 29.9 & 32.8 & 34.5 & 48.3 \\
 & More Information           & 45.2 & 8.2 & 46.6 & 20.6 & 38.2 & 41.2 & 17.2 & 55.2 \\
 & Consistency                & 84.9 & 1.4 & 13.7 & 35.3 & 29.4 & 35.3 & 37.9 & 51.7 \\
 & Linguistic Confidence      & 79.5 & 5.5 & 15.1 & 36.8 & 19.1 & 44.1 & 37.9 & 41.4 \\
\midrule
\multirow{4}{*}{\textsc{inst tune}}
 & \textit{Majority vote} & 86.3 & 4.1 & 9.6 & 38.9 & 28.3 & 32.8 & 24.1 & 41.3 \\
 & More Information           & 49.3 & 4.1 & 46.6 & 23.5 & 30.9 & 45.6 & 13.8 & 48.3 \\
 & Consistency                & 87.7 & 0.0 & 12.3 & 36.8 & 20.6 & 42.6 & 34.5 & 37.9 \\
 & Linguistic Confidence      & 79.5 & 4.1 & 16.4 & 35.3 & 26.5 & 38.2 & 24.1 & 48.3 \\
\midrule
\multirow{4}{*}{\textsc{pre inst tune}}
 & \textit{Majority vote} & 59.0 & 38.3 & 2.7 & 64.2 & 26.8 & 9.0 & 44.9 & 48.2 \\
 & More Information           & 54.8 & 27.4 & 17.8 & 66.2 & 26.5 & 7.4 & 24.1 & 55.2 \\
 & Consistency                & 54.8 & 34.2 & 11.0 & 61.8 & 23.5 & 14.7 & 51.7 & 41.4 \\
 & Linguistic Confidence      & 50.7 & 32.9 & 16.4 & 50.0 & 23.5 & 26.5 & 41.4 & 48.3 \\
\bottomrule
\end{tabular}%
}
\caption{Per-method breakdown of confidence estimators for LLaMA3.1-8B-Instruct on the Claim Judgment task. Each knowledge update method is evaluated with the individual confidence estimators and the majority vote combination. Values are percentages.}
\vspace*{4mm}
\end{table*}
 
The results show that the relative ordering of knowledge update methods remains consistent across all four confidence estimators, while in some cases the \textit{More Information} method might flag uncertainty more aggressively, inflating loss values. Overall, this indicates that our key findings are robust to the choice of confidence estimator, and that majority voting strikes a reasonable balance between calibration and stability.

\section{Scientific LLMs}
\label{sec:scientific_llms}

\begin{table*}[h!]
    \centering
    \setlength{\tabcolsep}{2pt}
    \resizebox{\textwidth}{!}{
    \begin{tabular}{lccc@{\hspace{6pt}}|ccc@{\hspace{6pt}}|cc} 
         \toprule[1.5pt]
          \textbf{Model} & 
          \textcolor{palevioletred}{\textbf{Preservation}} &
          \textcolor{palevioletred}{Distortion} &
          \textcolor{palevioletred}{Loss} &
          \textcolor{lightblue}{\textbf{Acquisition}} &
          \textcolor{lightblue}{Distortion} & 
          \textcolor{lightblue}{Loss} & 
          \textcolor{goldenrod}{\textbf{Projection}}& 
          \textcolor{goldenrod}{Loss}\\ 
          \midrule[0.75pt]
         \textsc{LLaMA3.1-8B-Instruct} & \textbf{72.0} & \underline{11.0} & \textbf{17.0} & \underline{15.5} & \textbf{13.4} & 71.7 & \textbf{16.8} & \underline{75.8}\\
         \textsc{OLMo2-7B} & 60.3 & 16.9 & \underline{22.8} & 11.9 & \underline{36.0} & \underline{52.1}& \underline{15.6} & 76.0\\
         \textsc{HoneyBee-7B} & \underline{62.6} & \textbf{0.0} & 37.4 & \textbf{18.2} & 42.8 & \textbf{39.0} & 15.1 & \textbf{69.0}\\
         \bottomrule[1.5pt]
    \end{tabular}
    }
    \caption{Performance of Standard Instruction-tuning on off-the-shelf and scientific LLMs in the claim judgment task within the domain of \emph{Materials Science}. Best results in \textbf{bold} and second best in \underline{underline}. 
    HoneyBee is a materials science model fine-tuned on LLaMa-7B.}
    \label{tab:scientific_llms}
    \vspace*{4mm}
\end{table*}

Rather than relying solely on off-the-shelf LLMs, researchers also use scientific LLMs~\citep{zhang2024comprehensive}, LLMs specifically trained or adapted for science. In this work, we also experiment with HoneyBee~\citep{song2023honeybee}, a llama-based model fine-tuned for \emph{Materials Science} domain using high-quality, relevant textual data from the open literature. As we find that the performance of scientific knowledge updates in \emph{Materials Science} is significantly lower than other domains (Section~\ref{sec:domains}), we wonder if applying on a specialized scientific LLM could help. As shown in Table~\ref{tab:scientific_llms}, scientific LLMs show no significant improvement compared to off-the-shelf LLMs of similar sizes, highlighting the unique challenges involved in updating scientific knowledge in terms of preservation, acquisition, and projection.

\section{Cross-domain Analysis}
\label{sec:cross-domain-analysis}
This section provides details on cross-domain analysis discussed in Section~\ref{sec:domains}.

\subsection{Citation Count}
See Table~\ref{tab:citation_domains} for the average citation counts of papers in our dataset across ten scientific domains, collected from Semantic Scholar~\citep{ammar2018construction}.

\begin{table}[h!]
\begin{center}
\setlength{\tabcolsep}{5pt}
\begin{tabular}{lc}
\toprule[1.5pt]
 \textbf{Domain} &\textbf{Citation Count}\\
\midrule[0.75pt]
Computer Science & 5.957 \\
Medicine & 4.575 \\
Biology & 5.799 \\
Materials Science & 7.192 \\
Psychology & 3.702 \\
Business & 2.447 \\
Political Science & 1.832 \\
Environmental Science & 5.973 \\
Agricultural and Food Sciences & 4.939 \\
Education & 2.002 \\
\bottomrule[1.5pt]
\end{tabular}
\end{center}
\caption{The average citation count of 1,000 conference or journal papers published between October 2022 and September 2023 across different domains.}
\label{tab:citation_domains}
\end{table}

\subsection{Domain-specific Tokens}
We randomly retrieve 1,000 conference or journal papers published between October 2022 and September 2023 for each of the ten domains. From these abstracts, we extract the 100 least frequently occurring tokens using the \textsc{LLaMA3.1} tokenizer. Stop words, punctuation, and numbers are removed from the list. The full list of tokens is provided in Table~\ref{tab:specialized_tokens_domains_1} and Table~\ref{tab:specialized_tokens_domains_2}.

\begin{table*}[h!]
    \centering
    \small
    \begin{adjustbox}{max width=1\textwidth}
    \begin{tabular}{lp{5in}}
         \toprule[1.5pt]
         {\textbf{Domain}} & \textbf{Specialized Tokens} \\
         \midrule[1pt]
        Computer Science
        &  [Background, chunks, mined, keywords, -res, ourced, Million, logistic, LR, encoder, coder, isolate, solved, imperfect, realized, abrupt, transmitted, connects, ended, tan, imoto, waveform, coefficients, -current, Self, -driving, navigation, drivers, orientation, camera, installed, videos, combinations, Next, Track, substantially, Thanks, mil, king, Trad, itionally, EB, -from, -more, including, dairy, deviations, sequent, lact, Est, herit, splitting, Rem, Mi, Together, encompass, setup, concluding, seaborn, matplotlib, Num, Py, Log, Literary, properly, client, -server, Client, Server, -Agent, Rapid, oring, planner, inverse, kin, ematic, fourth, Transfer, HTTP, send, Wireless, Control, missions, envi, nets, -agent, Path, -aware, preservation, extends, intermediate]\\ \midrule[0.75pt]
        Medicine
        &  [logic, outputs, AY, applic, subt, ropical, underscores, gam, publicly, overlapping, warrant, intimate, website, Domestic, Violence, DV, item, DV, observation, attainment, jun, foregoing, divorce, offspring, focused, uns, aturated, UF, palm, Animals, aily, Spatial, lost, -Jan, uary, Identification, Matrix, Laser, Ion, Time, Flight, rometer, MAL, Possible, encountering, opportun, Admission, inertia, RAP, charge, iny, -fe, alan, mem, brane, reversed, PAR, carbohydrate, -chain, aur, -en, rich, chicken, iated, recipients, misconception, abandonment, sustaining, optim, -ag, Enhanced, -k, Da, property, aiding, TJ, Adopt, -condition, polarization, Moh, -Tr, optic, interf, amil, arth, ref, dup, sister, ismus, disclosed, fe]\\ 
        \midrule[0.75pt]
        Biology
        &  [Qu, QS, attracts, basics, realization, solving, oriented, priorities, productive, oking, subt, timely, priority, ials, uns, aturated, Di, Twenty, palm, Animals, aily, -k, aiding, ulcer, rebound, TJ, Adopt, ada, Br, voltage, Nav, hurdle, arr, hyth, mic, :c, :p, Nav, exponentially, impose, UG, specialised, Laur, Material, -response, iaux, -comp, -death, Cock, ayne, olated, -rate, visceral, Unexpected, dc, Sp, pm, Nos, Statement, Kid, okes, emergency, monitored, urine, elo, album, -sk, ewed, Highly, inherently, paths, quasi, Americans, idi, opathic, ATIC, III, restrictive, Character, omorphic, stature, -height, Binding, slow, cognition, BF, doubled, unexpectedly, Well, come, Council, Horizon]\\ 
        \midrule[0.75pt]
        Materials Science
        &  [Even, afford, lacks, Rh, unnecessary, seem, believed, restrictive, transistor, NR, ampl, -terminal, stand, gate, COM, vertically, dissertation, satisfy, gien, Regular, convinced, rows, satin, stitch, ext, skipping, stitches, spent, row, Regression, duce, choose, passes, -cons, istent, VP, -mult, VP, supplemented, strengthened, deflect, meticulous, seems, Hamilton, energetic, warp, shr, mesh, widening, inferior, Spacer, twisting, earlier, oogeneous, alter, retained, vil, abundance, ol, clin, partition, Applying, minute, iles, Tit, -visible, follow, -first, law, Ev, Add, itive, Manufacturing, yx, ylon, chopped, Fil, Fabric, manners, inevitable, Increase, Rotary, sty, ABS, isot, chips, dispens, conforms, Cross, VR, incomplete]\\ 
        \midrule[0.75pt]
        Psychology
        &  [supporters, Turkish, Federation, league, season, mekte, aca, reside, deploy, February, undergone, referral, .Pre, vious, Using, partner, .Actor, .Al, though, careers, disabling, hab, ilitation, vo, rehabilit, ROT, Reality, pencil, Compar, ventional, uo, -exec, expanding, Council, arts, theatrical, Documentation, encode, professor, Yuri, Kon, stant, ovich, orn, twenties, ungen, Menschen, affen, Ap, ensured, Wall, Thought, Anyway, -long, Sol, wind, -al, gorithm, izable, Simon, scientists, .L, Rub, Brush, insky, Ya, onom, contempor, Kor, la, Login, .F, Spi, rid, Lap, isto, pol, sk, .N, Sav, ols, outlines, specializing, Today, attracted, squares, phys, immense, scan, ancers, dancers, Oper, recognizing]\\ 
    \bottomrule[1.5pt]
    \end{tabular}
    \end{adjustbox}
            \caption{Specialized tokens by domains.} 
    \label{tab:specialized_tokens_domains_1}
    \vspace*{-2mm}
\end{table*}

\begin{table*}[h!]
    \centering
    \small
    \begin{adjustbox}{max width=1\textwidth}
    \begin{tabular}{lp{5in}}
         \toprule[1.5pt]
         {\textbf{Domain}} & \textbf{Specialized Tokens} \\
         \midrule[1pt]
        Business
        &  [universities, shaped, intents, Planned, variance, cur, ricula, colleges, hiding, thinks, Mixed, edir, esign, Ped, est, rian, busiest, worship, Aut, ad, plan, Sketch, stone, lamps, disabilities, night, shade, trees, trash, cans, benches, ender, neutrality, pose, proves, Cycling, territory, initially, loc, quali, quant, -line, Content, ardin, -art, supervised, ervised, Random, Boost, IW, AL, CH, AG, PH, tactics, Analy, engaging, Among, tips, istrict, awi, Boston, Consulting, regulator, backward, penetration, automobiles, -side, Rating, internal, insurance, distribution, Observation, interpreted, bands, Merch, andise, band, tok, po, plain, publish, stories, inders, keepers, consent, Customers, annoyed, technology, ynamic, breakdown, Revenue, GRA, Tam]\\ 
        \midrule[0.75pt]
        Political Science
        &  [Mayor, alignments, reputation, cular, tapping, -period, Deputy, Chair, chairman, upcoming, Glob, unexpected, bur, sts, -demand, ding, negligence, -ray, ultrasound, oxygen, cylinders, bribery, coll, usion, cov, -care, India, rebuild, aped, reproductive, Pregnancy, Assessment, Monitoring, merged, -unit, corresponds, predicted, Models, Medicaid, uninsured, constr, lethal, essential, inaccurate, hypothetical, underestimate, innocent, ale, assass, massac, adopts, rig, idity, Usage, Use, Sig, Received, affili, omics, Highlands, Ranch, Colorado, Ang, lia, Norfolk, Economics, Biology, eos, Cor, respond, Andrew, Page, prosper, chaotic, breaks, Reports, Officials, slap, scr, ulous, collusion, receives, update, unify, Method, -trans, subordinate, ordination, prescribed, abandon, status, liquid, ields]\\ 
        \midrule[0.75pt]
        Environmental Science
        &  [Ach, arya, Narendra, Technology, Kum, anj, Ay, hya, .P, ban, horizontally, Sm, breaks, combust, ibles, -contained, breathing, charger, differentiated, gar, mist, charged, firefighters, etal, attractive, afford, hollow, template, lacks, alloy, aceous, giving, mo, ieties, Associated, flatt, aling, current, illustrates, Growing, arms, easing, offsets, margins, sentinel, Gui, Woody, Native, Increase, ensured, Gas, economical, ENT, uction, pression, -dis, charge, formula, Fresh, someone, wants, easiest, acronym, Add, Assessment, Alternative, opted, executed, adversely, effected, Jas, Percent, retain, igated, executor, affairs, ochrome, ringing, anch, Geo, -grid, PL, Net, rein, forc, -ing, Mon, omantic, trace, retained, diamond, vil, syn, analog]\\ 
        \midrule[0.75pt]
        \makecell[l]{Agricultural and\\Food Sciences}
        &  [Background, arms, easing, -offs, offsets, sentinel, Gui, Version, Woody, Native, Imp, Increase, Trees, expense, Across, que, stration, Ins, bodily, Large, Blue, elle, Wood, Color, guaranteed, components, vit, dispersion, mixer, completeness, sustaining, urgently, scheduling, intric, sovereignty, expenditures, spending, excessively, pleasing, purple, Jerusalem, Hel, thus, Cal, brom, igh, yer, Fiber, Analyzer, -An, kom, zap, mango, Sit, aja, Sap, vene, rys, llum, lance, andra, -J, reserves, subsets, categorized, identical, fairness, CNN, impressive, showcasing, Bihar, consequ, odule, attrib, inferred, unavoidable, worrying, conscient, happens, prescribed, impossible, shed, anticipation, CCC, cricket, WF, blends, CCC, Purchase, atisfied, usted, assert, predictors, Leipzig, Actual, fuels]\\ 
        \midrule[0.75pt]
        Education
        &  [Jur, udence, Enough, Weak, magnitude, =a, +b, RTL, param, etric, Plans, Infrastructure, Super, Japanese, Young, intents, Planned, squares, legitimate, recognized, launch, Br, song, gains, core, ivism, Ecology, Human, entity, instantly, environmentally, Ec, teeth, minimized, .Result, .Con, waves, alyze, Evaluate, WAR, PER, IOD, ropy, yz, hev, Regional, Archive, rad, martial, informational, histor, resist, battlefield, acting, aters, Ukrain, protect, fitting, super, asks, otic, Lim, Tang, gam, overlapping, ineffective, Liter, Connected, compat, Evalu, SET, unten, redesign, -made, checked, weighted, Messenger, emails, iber, rooted, emancip, deficit, implicitly, depr, overly, applic, -created, Ps, LD, omencl, etiquette, ingu]\\ 
    \bottomrule[1.5pt]
    \end{tabular}
    \end{adjustbox}
            \caption{Specialized tokens by domains. (Continued)} 
    \label{tab:specialized_tokens_domains_2}
\end{table*}

\subsection{Occurrence in Pretraining Corpus}
Refer to Table~\ref{tab:pretraining_occurrence_domains} for the average occurrence of domain-specific tokens in the Dolma-v1.7 pretraining corpus~\citep{dolma}, collected using Infini-gram~\citep{Liu2024InfiniGram}.

\begin{table*}[h!]
\begin{center}
\vspace*{2mm}
\setlength{\tabcolsep}{5pt}
\begin{tabular}{lc}
\toprule[1.5pt]
 \textbf{Domain} &\textbf{Average Token Occurrence}\\
\midrule[0.75pt]
Computer Science & 32966797 \\
Medicine & 33036396 \\
Biology & 30569548 \\
Materials Science & 39959970 \\
Psychology & 34891007 \\
Business & 42227384 \\
Political Science & 24943232 \\
Environmental Science & 32928017 \\
Agricultural and Food Sciences & 20853024 \\
Education & 27514910 \\
\bottomrule[1.5pt]
\end{tabular}
\end{center}
\vspace{-2mm}
\caption{The average number of occurrence of domain-specific tokens (as identified in Table~\ref{tab:specialized_tokens_domains_1} and Table~\ref{tab:specialized_tokens_domains_2}) in the Dolma-v1.7~\citep{dolma} pretraining corpus.}
\label{tab:pretraining_occurrence_domains}
\end{table*}

\subsection{Correlation}
\begin{figure*}[h!]
    \centering
    \includegraphics[width=1.0\textwidth]{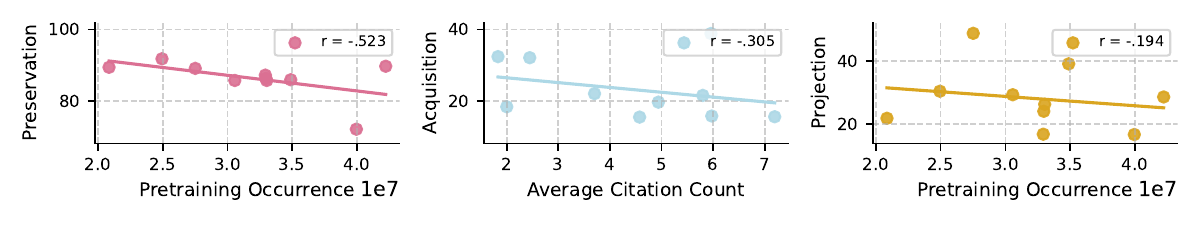}
    \vspace*{-8mm}
    \caption{The correlation between \textcolor{palevioletred}{Preservation}, \textcolor{lightblue}{Acquisition}, \textcolor{goldenrod}{Projection} and average citation count, as well as pretraining occurrence. The closer the points are to the best‑fit line, the stronger the correlation.}
    \label{fig:domain-factors-2}
\end{figure*}

Figure~\ref{fig:domain-factors-2} presents the correlation between Preservation and pretraining occurrence, Acquisition and average citation count, as well as Projection and pretraining occurrence.

\section{Discussion and Justification}


\subsection{Knowledge Projection}
\label{sec:discussion_knowledge_projection}
We use claim classification and claim generation as a practical way to operationalize and quantify aspects of future scientific knowledge. We view it as a tractable and meaningful approximation that allows controlled evaluation. Our goal is to assess whether a model can use its internalized knowledge to form a reasonable sense of potential future discoveries. The ability to extrapolate from existing knowledge, recognize emerging patterns, and generalize to future findings is an important aspect of forecasting in scientific contexts. Our evaluation measures whether models can anticipate or correctly judge claims that reflect subsequent scientific developments, which we consider a meaningful and measurable facet of forecasting ability.

We define Knowledge Projection as the percentage of claims in $\mathcal{P}_\text{future}$ that transition from \textit{unknown} to \textit{correct} after the knowledge update. Two conditions must be satisfied:
\begin{itemize}
    \item The claim is unknown before the update.
    \item The claim is correct with high confidence after the update.
\end{itemize}

In other words, prior to the knowledge update, the model is unable to answer the claim with sufficient confidence (regardless of whether its low-confidence response happens to be correct or incorrect). After the update, the model answers correctly and with high confidence. The only variable that changes between these two responses is the knowledge update itself. This controlled setup isolates the effect of the introduced knowledge trajectory. Therefore, any systematic shift from ``unknown'' to ``correct with high confidence'' can be attributed to the knowledge update, rather than to domain-wide priors or generic scientific templates. While some stochastic variation is inevitable in generative models, the controlled before–after comparison provides direct evidence that the projection is conditioned on the introduced knowledge. If the model were merely relying on field-level expectations, we would not expect such transition.

\subsection{Other Models}
\label{sec:other-models}
We additionally evaluate two more models: OLMo-2-7B-Instruct and Llama-3.2-3B-Instruct, using standard instruction-tuning and inference settings, respectively. Results (Table~\ref{tab:additional_models}) on these additional smaller models further support our conclusion: smaller models generally require training, and inference-based methods alone are often insufficient due to their limited in-context learning capabilities.

\begin{table*}[h]
\centering
\small
\resizebox{\linewidth}{!}{
\begin{tabular}{lccc|ccc|cc}
\toprule[1.5pt]
\textbf{Method} & \textcolor{palevioletred}{\textbf{Preservation}} &
          \textcolor{palevioletred}{Distortion} &
          \textcolor{palevioletred}{Loss} &
          \textcolor{lightblue}{\textbf{Acquisition}} &
          \textcolor{lightblue}{Distortion} & 
          \textcolor{lightblue}{Loss} & 
          \textcolor{goldenrod}{\textbf{Projection}}& 
          \textcolor{goldenrod}{Loss}\\
\midrule[1pt]
\multicolumn{9}{c}{\textsc{\textbf{Llama-3.2-3B-Instruct}}}\\ 
\midrule[1pt]
 \textcolor{black}{\textsc{inst tune}}& 56.2 & 21.0 & 22.8 & 29.5 & 29.6 & 40.9 & 23.7 & 30.2 \\
\textcolor{black}{\textsc{infer}} & 51.3 & 35.6 & 13.1 & 30.0 & 31.3 & 38.7 & 25.2 & 18.8 \\
\midrule[1pt]
\multicolumn{9}{c}{\textsc{\textbf{OLMo-2-7B-Instruct}}}\\ 
\midrule[1pt]
\textcolor{black}{\textsc{inst tune}} & 72.0 & 8.9 & 19.1 & 37.6 & 31.6 & 30.8 & 25.5 & 40.2 \\
\textcolor{black}{\textsc{infer}} & 62.2 & 21.9 & 15.9 & 35.2 & 41.0 & 23.8 & 29.3 & 24.0 \\
\bottomrule[1.5pt]
\end{tabular}
}
\caption{Results on additional models under claim judgment task.}
\label{tab:additional_models}
\end{table*}


\subsection{Abstract instead of Full paper}
\label{sec:discussion_abstract}
We use abstracts as the primary source of scientific content for the following reasons:

\begin{itemize}
    \item \textbf{Copyright and data access.} Using the full text of papers can introduce practical constraints related to data availability and copyright restrictions, which limit large-scale, systematic use of full articles.
    
    \item \textbf{Context window limitations.} Incorporating full papers into the context window is challenging, particularly for the model sizes considered in this work. We also did not rely on closed commercial models for this analysis, as their knowledge cutoffs are not transparent and they do not allow training or controlled updating, which is necessary for our experimental setting.
    
    \item \textbf{Consistency with prior work.} Our approach follows prior research in scientific domains that also uses abstracts as the primary source of scientific content \citep{wadden2020fact, newman2024arxivdigestables}.
\end{itemize}

\subsection{Grouping Scientific Knowledge with Citations}
\label{sec:discussion_group_knowledge}
Citation links are the signal we use to group knowledge and we want to argue that scientific progress is usually not a strict continuation of a single, narrowly defined research topic. Using the example of transformers, given a specific topic, what evolves most frequently and rapidly is often not a direct extension of the core architecture itself, but related ideas, such as new applications of the architecture, analyses of its properties, or methodological variations centered around the same general area. Direct follow-up work that proposes entirely new architectures certainly exists, but it tends to be less common than these broader, related developments. The broader developments are often the actual driving force behind subsequent direct advances.

Our goal is to mimic real scientific progress in the wild, which is often more scattered and heterogeneous than a sequence of tightly connected, linear continuations. Accordingly, when we update a model with new scientific progress, we aim to evaluate to what extent the model can preserve prior knowledge, acquire new knowledge, and project forward. This projection does not necessarily have to be a direct continuation of a specific topic; it may also involve generating or reasoning about work that is broadly related within the surrounding research area. Closely related knowledge is indeed more likely to be impacted, but broader knowledge may also be reshaped, forgotten, blurred, or recontextualized as scientific understanding evolves. Importantly, we do not discard directly related knowledge. Such work is naturally included within the subset of cited papers and therefore remains part of our evaluation. Our approach extends the scope to additionally measure influence on more generally related work, aiming for a more comprehensive assessment of knowledge evolution.

\subsection{Scalability}
Our contribution is primarily the methodology and evaluation framework, rather than a fixed dataset.
We have validated the effectiveness of our data curation process, particularly the claim generation methods, which makes the framework adaptable to new domains or to models with different temporal knowledge cutoffs. In practice, adapting the framework only requires collecting relevant papers and sampling new claims, rather than redesigning the evaluation from scratch.
Moreover, if the difference between knowledge cutoff dates is not substantial, users or practitioners can also reuse appropriate subsets of the existing dataset, which further reduces reconstruction costs and improves scalability.

\end{document}